\documentclass[manuscript]{acmart}

\AtBeginDocument{%
  \providecommand\BibTeX{{%
    \normalfont B\kern-0.5em{\scshape i\kern-0.25em b}\kern-0.8em\TeX}}}
\usepackage{ulem}
\usepackage{multirow}
\usepackage{graphicx}
\usepackage{wrapfig}
\usepackage{subfigure}
\usepackage{bbding}
\def\model{TAR}

\acmISBN{978-1-4503-XXXX-X/18/06}

\begin{document}

\title{Topology-aware Reinforcement Feature Space Reconstruction for Graph Data}


\author{Wangyang Ying}
\orcid{0009-0009-6196-0287}
\affiliation{
  \institution{Arizona State University, School of Computing and Augmented Intelligence, Tempe}
  \country{USA}}
\email{yingwangyang@gmail.com}

\author{Haoyue Bai}
\orcid{0009-0009-1328-9230}
\affiliation{%
  \institution{Arizona State University, School of Computing and Augmented Intelligence, Tempe}
  \country{USA}
}\email{baihaoyue621@gmail.com}

\author{Kunpeng Liu}
\orcid{0000-0002-6053-5977}
\affiliation{%
 \institution{Portland State University, Department of Computer Science, Portland}
 \country{USA}}
\email{kunpeng@pdx.edu}

\author{Yanjie Fu\textsuperscript{\textdagger}}
\orcid{0000-0002-1767-8024}
\affiliation{%
  \institution{Arizona State University, School of Computing and Augmented Intelligence, Tempe}
  \country{USA}}
\email{yanjie.fu@asu.edu}
\thanks{\textsuperscript{\textdagger} Corresponding Author}
\renewcommand{\shortauthors}{Ying et al.}

\begin{abstract}
    Feature space is an environment where data points are vectorized to represent the original dataset. Reconstructing a good feature space is essential to augment the AI power of data, improve model generalization, and increase the availability of downstream ML models. 
    Existing literature, such as feature transformation and feature selection, is labor-intensive (e.g., heavy reliance on empirical experience) and mostly designed for tabular data. Moreover, these methods regard data samples as independent, which ignores the unique topological structure when applied to graph data, thus resulting in a suboptimal reconstruction feature space. Can we consider the topological information to automatically reconstruct feature space for graph data without heavy experiential knowledge?
    To fill this gap, we leverage topology-aware reinforcement learning to automate and optimize feature space reconstruction for graph data. Our approach combines the extraction of core subgraphs to capture essential structural information with a graph neural network (GNN) to encode topological features and reduce computing complexity. Then we introduce three reinforcement agents within a hierarchical structure to systematically generate meaningful features through an iterative process, effectively reconstructing the feature space.
    This framework provides a principled solution for attributed graph feature space reconstruction. The extensive experiments demonstrate the effectiveness and efficiency of including topological awareness. Our code and data are available at \url{https://tinyurl.com/graphFT123}.

\end{abstract}



\setcopyright{acmlicensed}
\acmJournal{TKDD}
\acmYear{2024}
\acmDOI{XXXXXXX.XXXXXXX}



\maketitle

\section{Introduction}

Artificial Intelligence (AI) development typically involves: 1) collecting data, 2) computing data representation (a.k.a., feature space),
and  3) applying machine learning (ML) models. In real world practices and deployments, the success of ML models highly relies on the quality of a feature space~\cite{representation}.  Developing an effective feature space is essential because it can reconstruct distance measures, reshape discriminative patterns, and augment the AI readiness (e.g., structural, predictive, interactions, and expression level) of data. 

\noindent 
Among a variety of data types (e.g., tabular, spatial, temporal, time series, sequence data), graph data exhibit the indispensable ability to capture interconnected and complex relationships, handle non-Euclidean spaces, facilitate information propagation, advance diverse domains (e.g., transportation systems, power grids, social networks, recommendation systems) in a way that traditional data types cannot. 
We study the problem of feature space reconstruction for attributed graph data. 
Formally, given an attribute graph described by a node-attribute matrix and an adjacent matrix, prediction labels, and a downstream ML task (e.g., node classification, link prediction, and graph classification), the objective is to automatically transform and reconstruct an optimal explicit feature space for the ML task.

\noindent Prior literature only partially addressed the problem:
1) classic feature engineering techniques, such as, empirical data preprocessing, feature selection~\cite{featureSelection1,featureSelection2,wangyang@fs,wang2024knockoffguidedfeatureselectionsingle,gong2024neurosymbolic}, and feature generation~\cite{featureGeneration1,featureGeneration2,wangyang@uft} are essential but labor-intensive. They can't achieve full automation in graph feature space reconstruction.  
2) recent automated feature space reconstruction methods~\cite{uddin2021pca,LDA,AFAT,NFS,TTG} are mostly designed for generic tabular data. When applied to graph data, they fail to consider the impact of topology structures of attributed graphs in building a graph feature space. 
3) neural graph representations methods~\cite{kipf2016semi,hamilton2017inductive,velickovic2017graph}, such as graph convolutional networks, graph attention networks, learn latent embedding feature representation without knowing attribute/feature names.
An interesting problem arises: can we develop an effective, automated explicit feature space reconstruction method for attributed graph data?

\noindent\textbf{Our insights: a topology-aware reinforcement generative perspective.} 
Our insights are two fold:
1) \uline{reinforcement generation}: 
Attributed graph feature space reconstruction can be viewed as an iterative feature cross-based generation decision process.  The feature knowledge of attributed graph data can be formulated as machine-learnable policies within a Reinforcement Learning (RL) framework. We show that 1) hierarchical RL can iteratively self-learn feature knowledge as policies to generate meaningful attributed graph features; 2) feature group-group crossing to generate large amounts of features can amplify reinforcement reward signals and accelerate feature space exploration.  
2) \uline{graph topology awareness}: there are two key elements in attributed graphs: (i) node attributes and (ii) graph topology. A native solution is to ignore graph topology and simplify attributed graphs as node attribute tabular data, then apply classic tabular data feature transformation methods. 
However, graph topology can't be ignored, because the great success of graph neural networks (GNNs) shows the importance of modeling neighborhood information and message passing and aggregation. 
We leverage graph topology from two angles: 
(a) large graphs with millions of nodes, exponentially increase computational costs. Topology awareness can help to focus on representative and frequent subgraphs, eliminate noisy nodes, and reduce computational costs. 
(b) RL needs to perceive the dynamic state updates of an attributed graph during iterative feature reconstruction. Topology awareness allows us to quantify the accurate state representation of an attribute graph via GNNs. 

\noindent\textbf{Summary of technical solutions.} Inspired by these insights, we propose a topology-aware reinforcement generative feature construction framework for attributed graphs. 
This framework has two goals: 1) effectively leveraging attributed graph topology; 2) automatically generating the optimal feature set from node attributes for a downstream ML task. 
To achieve Goal 1, we propose to mine core subgraphs of the dataset to capture the important and valuable local structure information. Subsequently, we use a GNN to learn the embedding of the subgraph as inputs to RL.
To achieve Goal 2, we design three reinforcement agents to automatically decide how to perform feature group-group crossing-based generation: a head feature agent to select the first feature group, an operation agent to select an operation (e.g., +, -, *), and a tail feature agent to select the second feature. We develop a hierarchical structure to coordinate agents to share states and learn better generation policies.

\noindent\textbf{Our contributions.} 1) \uline{Perspective}:  We formulate an important task: attributed graph feature space reconstruction, from a topology-aware reinforcement generative perspective. 2) \uline{Framework}: We develop a subgraph-grouping-reinforcement framework. 3) \uline{Computing}:  our idea of spotting core subgraphs as original graph approximates can reduce costs without losing accuracy on graph feature transformation; our strategy of using graph neural networks as reinforcement state quantifiers can accurately perceive the state change of attributed graph feature space; our solution of hierarchical agent structure and group-level feature crossing can augment reward feedback, speed up exploration,  learn better graph feature knowledge policies for automation.

\section{Definitions and Problem Statement}

\subsection{Important Definitions}

\noindent{\textbf{Feature Cross.}} We aim to reconstruct feature space for an attributed graph dataset $\mathcal{G} = (\mathbf{A}, \mathbf{X}, y)$. Here, $A$ is an adjacency matrix, $\mathbf{X} = [\mathbf{f}_1, \mathbf{f}_2, ..., \mathbf{f}_n]$ is a feature matrix, in which a row is a node and a column is a feature, and $y$ is the target labels of nodes or edges. 
We apply mathematical operations to transform the original features of the attributed graph into new features (e.g., $[\mathbf{f}_1, \mathbf{f}_2] \rightarrow [sin(\mathbf{f}_1), \mathbf{f}_1 + \mathbf{f}_2]$). 
We employ two types of operations in our proposed method: 1) unary operations including`$log, exp, sin$', etc; 2) binary operations including `$+, -, *, /$'.  

\subsection{Problem Statement}
Given a graph dataset $\mathcal{G} = (\mathbf{A}, \mathbf{X}, y)$ that includes the adjacency matrix $\mathbf{A}$, the feature matrix $\mathbf{X}$, and the target labels $y$. 
Our goal is to automatically reconstruct an optimal and explicit feature space $\mathbf{X}^*$ for the attributed graph in order to improve a downstream ML task $\mathcal{M}$ (e.g., node classifications, link predictions, or graph classifications. For instance, when $\mathcal{M}$ is link prediction, $y$ indicates whether any of two nodes are connected). 
The objective is given by:
\begin{equation}
    \mathbf{X}^* = \arg\max_{\hat{\mathbf{X}}} (\mathcal{V}_{\mathcal{M}}(\hat{\mathbf{X}}, y)),
\end{equation}
where $\hat{\mathbf{X}}$ can be viewed as a feature subset that includes multiple graph original features and feature crosses (e.g., $[\mathbf{f}_1, \mathbf{f}_3, sin(\mathbf{f}_1), ..., \mathbf{f}_1 + \mathbf{f}_2]$), and $\mathcal{V}$ is the predictive performance indicator.
\section{Attributed Graph Feature Transformation Learning}

\subsection{Overview of Proposed Framework}

\begin{figure}
    \centering
    \includegraphics[width=1\textwidth]{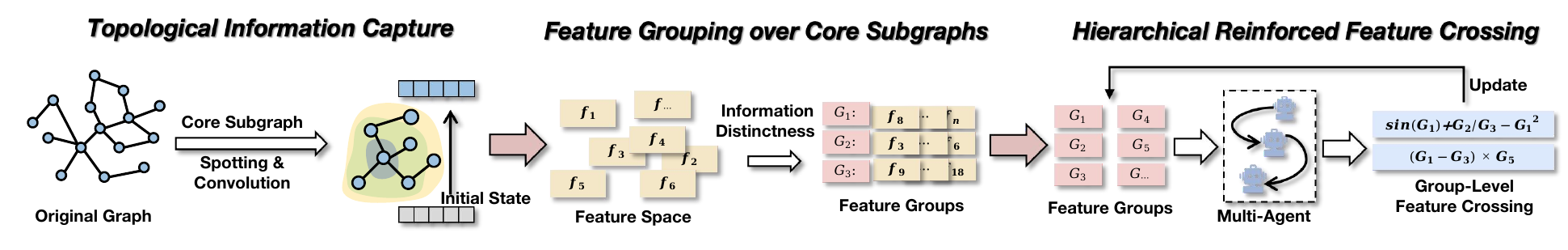}
    \caption{An Overview of {\model}. First, we spot the core subgraph from the original dataset and embed it into a fixed-length embedding to capture the topological information. Second, we group the attributes (a.k.a, features) of the subgraph to enrich the information of features. Finally, we develop a reinforcement system to select features and operations to reconstruct the graph feature space.}
    \label{framework}
\end{figure}
Figure~\ref{framework} illustrates the \textbf{T}opological-\textbf{A}ware \textbf{R}einforcement (\textbf{TAR}) feature space reconstruction framework includes three steps. 
1) \textbf{Topological Information Capture with Core Subgraphs and GNNs;}
2) \textbf{Graph Feature Grouping for Amplifying Reinforcement Signals;}
3) \textbf{Hierarchical Reinforced Feature Crossing.}
Specifically, in phase 1, we mine core subgraphs from the original graph, and then we devise a graph neural network to embed the subgraph into a fixed-length embedding vector as inputs to capture the topological information.
In phase 2, we cluster the original features (node attributes) into different feature groups by minimizing intra-group feature redundancy, and maximizing inter-group feature distinctness. 
In phase 3, we develop reinforcement feature crossing between groups to generate multiple features at each time and add generated features to the original feature set. In particular, we develop a hierarchical reinforcement learning method to learn three agents to select the two most informative feature groups and the most appropriate operation. We regard a downstream performance of the reconstructed feature space as a reward, and leverage the Bellman equation~\cite{bellman} to optimize the agents to learn the optimal policies.

\begin{figure}
    \centering
    \includegraphics[width=0.8\textwidth]{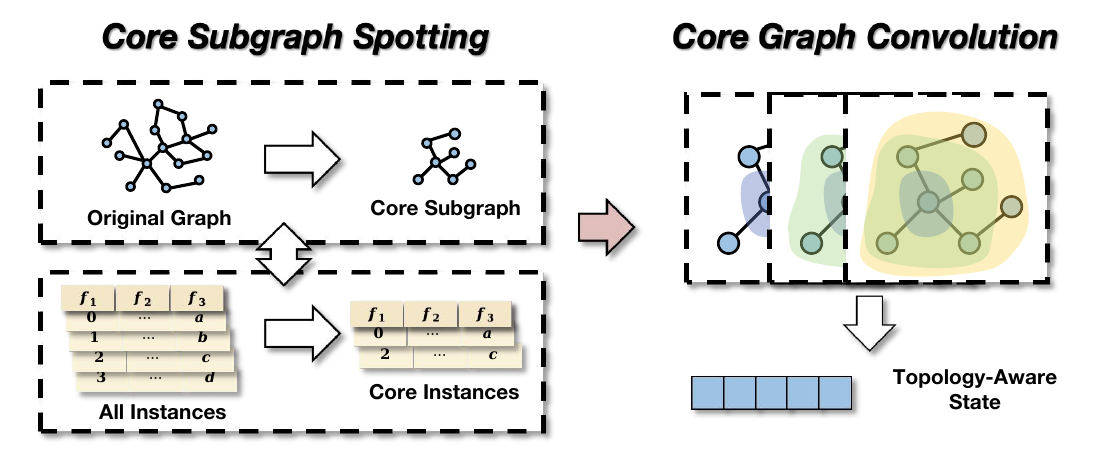}
    \caption{Topological Information Capture with Core Subgraphs and GNNs.}
    \label{subgraph}
\end{figure}
\subsection{Topological Information Capture with Core Subgraphs and GNNs}
\label{GNN-state}
\textbf{\textit{1) Reshaping Representation: Core Subgraphs as Original Graph Approximations for Fast Attributed Graph Feature Transformation.}}

\noindent\textbf{Why mining frequent subgraphs matters?}
Frequent subgraphs refer to subgraphs that frequently occur in a given graph, and represent the important relational topology, structures, and characteristics of the graph.
We propose a subgraph-based rectified feature-crossing strategy, by mining frequent subgraphs of an attributed graph, thereafter, conducting feature crossing over the node features of frequent subgraphs instead of entire graphs. 
This strategy can advance
1) \textit{\uline{transformation efficiency:}}  reduce the number of nodes that participated in attributed graph feature transformation, and reduce computational costs;
2) \textit{\uline{information quality:}}  remove noise,  biased, border nodes,  retain core and informative graph structures, and model the most intrinsic graph topology for feature transformation. 

\noindent\textbf{Spotting core subgraphs.}
Inspired by~\cite{gSpan}, we employ an efficient and accurate algorithm to identify core subgraphs in the training data, as shown in Figure~\ref{subgraph}. This algorithm utilizes the 
Depth-First Search (DFS) to explore subgraphs:
1) Starting from each node to explore subgraphs;
2) progressively expanding subgraphs while computing the support values (the number of occurrences in the dataset).
We assign a unique code to each explored path by DFS, representing a subgraph. This strategy can effectively enumerate all core subgraphs to reduce the complexity of the dataset. 

\noindent\textbf{\textit{2) Embedding Core Subgraphs: GNN Aggregation for Fixed-Length Graph Representations to Capture Topological Information.}}

\noindent\textbf{Why embedding core subgraphs matters?} Aggregating subgraph representations is essential for capturing the core topological and attribute information of a graph in a way that is both compact and meaningful. By embedding core subgraphs into a fixed-length vector representation, we provide a rich, structured input for reinforcement learning (RL) models. This representation allows the RL agent to leverage both local and global graph information, enabling more effective learning and decision-making. Specifically, the GNN-generated vector captures key structural patterns and interrelationships within the graph, making it an ideal input for RL algorithms that require a consistent and informative state representation. The aggregated vector representations retain the core structural information of subgraphs, enabling the RL model to focus on topological patterns. In essence, this approach combines the strengths of GNNs for encoding graph structure and RL for optimizing decisions based on these embeddings, ultimately leading to more effective solutions for graph feature transformation problems.

\noindent\textbf{Aggregating Core Subgraphs.} To capture the structure information inherent in graphs, we use a pre-trained GNN to embed a graph into a fixed-length vector as input. Specifically, given a node $v$ from the graph, we sample its all neighbor nodes $\{u_1, u_2, ..., u_p\}$, where $p$ is the number of neighbors. The $\mathbf{h}_v$ and $\mathbf{h}_u$ are the node representations of node $v$ and node $u$ respectively. We aggregate the representation of all neighbors by calculating their mean: 
\begin{equation}
\overline{\mathbf{h}_u} = \frac{1}{p}\sum_{i=1}^p\mathbf{h}_{u_i}
\end{equation}
We update the representations of node $v$ by:
\begin{equation}
\mathbf{h}_v = ReLU(\mathbf{W} \cdot CONCAT(\mathbf{h}_v, \overline{\mathbf{h}_u})),
\end{equation}
where ReLU is the activation function and $\mathbf{W}$ is the well-trained weight matrix. Finally, we average all node representations to extract the graph-level representation and regard it as the input of RL.

\subsection{Feature Grouping over Core Subgraphs}
\begin{figure}
    \centering
    \includegraphics[width=1.0\textwidth]{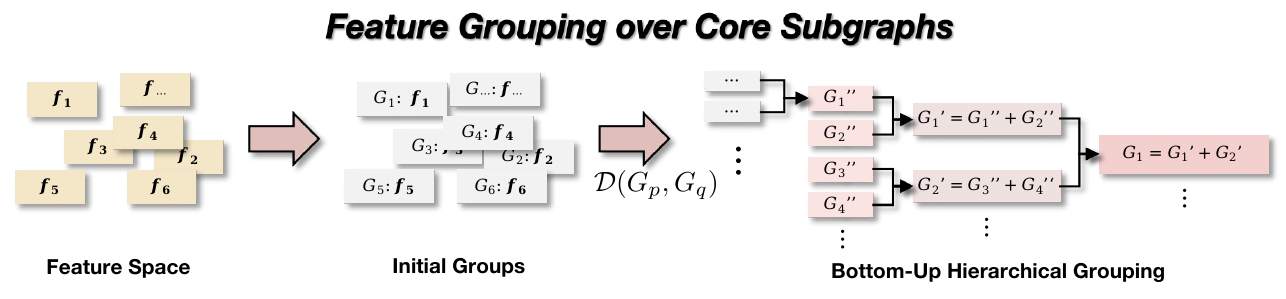}
    \caption{Feature Grouping over Core Subgraphs. We group the graph attributes (a.k.a., features) using bottom-up hierarchical grouping evaluated by group information distinctness.}
    \label{part_2}
\end{figure}
\textbf{Why grouping features matters?} 
We propose a strategy to accelerate the exploration and crossing of features; that is, firstly group the features of core subgraph nodes and then perform feature crossing between two feature groups, as opposed to crossing two individual features. 
In this way, we can generate more crossed features instead of one crossed feature, provide stronger reward feedback, and incentivize agents to update policy learning faster. 
Another insight is that diversity plays a role in feature crossing. In other words, in a feature space, if the features being crossed exhibit significant information distinctness, it is more likely to generate meaningful new features for augmenting data representation; on the other hand, if the features being crossed are similar, the generated features have low information gain. 
Based on these two principles, we propose to group subgraph features by information distinctiveness, followed by crossing the feature groups. The two feature groups should satisfy the following conditions: 1) the information between each feature group and the predictive label $y$ should show significant differences; otherwise, the two groups should be further merged; 2) the redundancy between the two feature groups should remain low.

\noindent\textbf{Quantify feature group-group information distinctness.} 
Given a node attribute matrix $\mathbf{X} = [\mathbf{f}_1, \mathbf{f}_2, ..., \mathbf{f}_n]$ of a mined subgraph and target labels $y$, we quantify feature group-group information distinctness by leveraging mutual information: 

\noindent\uline{1) Relevance between a feature group and target labels.} Formally, given a feature group $G_p = [\mathbf{f}_1, \mathbf{f}_3,...,\mathbf{f}_p]$ (a subset of the node attribute matrix $\mathbf{X}$), we calculate the relevance by:
\begin{equation}
    \mathcal{R}el(G_p, y) = \frac{1}{|G_p|} \sum_{\mathbf{f}_i \in G} \mathbb{I}(\mathbf{f}_i; y),
\end{equation}
where $\mathbb{I}(\cdot;\cdot)$ denotes mutual information and $|G_p|$ is the number of features in group.

\noindent\uline{2) Redundancy between two feature groups.} Formally, given two feature groups $G_p = [\mathbf{f}_1, \mathbf{f}_3,...,\mathbf{f}_p], G_q = [\mathbf{f}_2, \mathbf{f}_5,...,\mathbf{f}_q]$, we calculate the redundancy between two feature groups by:
\begin{equation}
    \mathcal{R}ed(G_p, G_q) = \frac{1}{|G_p|\cdot|G_q|} \sum_{\mathbf{f}_i \in G_p, \mathbf{f}_j \in G_q}\mathbb{I}(\mathbf{f}_i;\mathbf{f}_j).
\end{equation}

\noindent\uline{3) Group information distinctness.} We construct a new evaluation metric to quantify the information distinctness between feature groups:
\begin{equation}
    \mathcal{D}(G_p, G_q) = \frac{|\mathcal{R}el(G_p, y) - \mathcal{R}el(G_q, y)|}{\mathcal{R}ed(G_p, G_q)+\delta},
\label{distinctness}
\end{equation}
where $\delta$ is a small positive constant to avoid division by zero errors. We use this evaluation metric to evaluate whether to merge $G_p$ and $G_q$. If the group information distinctness $\mathcal{D}(G_p, G_q)$ is very low, which means we should put these two group features together. 

\noindent\textbf{Bottom-up node feature grouping of attributed graphs.} 
We leverage a hierarchical bottom-up clustering approach to create small groups first and then combine small groups with low inter-group feature information distinctness into large groups. Our method includes two steps: 1) the initialization stage; 2) the bottom-up merging stage. 
Specifically, in the initialization stage, given a core subgraph, we use $\mathbf{X}$ to represent the node feature matrix, with each row as a node and each column as a feature. We denote $\mathbf{X}$ as a set of feature vectors $[\mathbf{f}_1, \mathbf{f}_2, ..., \mathbf{f}_n]$. We regard each single feature as a small feature group. In the merging stage, we iterate the following steps: i) calculate the \textbf{group information distinctness $\mathcal{D}$} as described in equation~\ref{distinctness} between any two groups, then merge the two groups with the lowest group information distinctness; ii) repeat i) until either the group information distinctness between all feature groups exceeds a predefined threshold or the number of feature groups reaches the minimum number of groups.

\subsection{Hierarchical Reinforced Feature Crossing}

\begin{figure}
    \centering
    \includegraphics[width=1.0\textwidth]{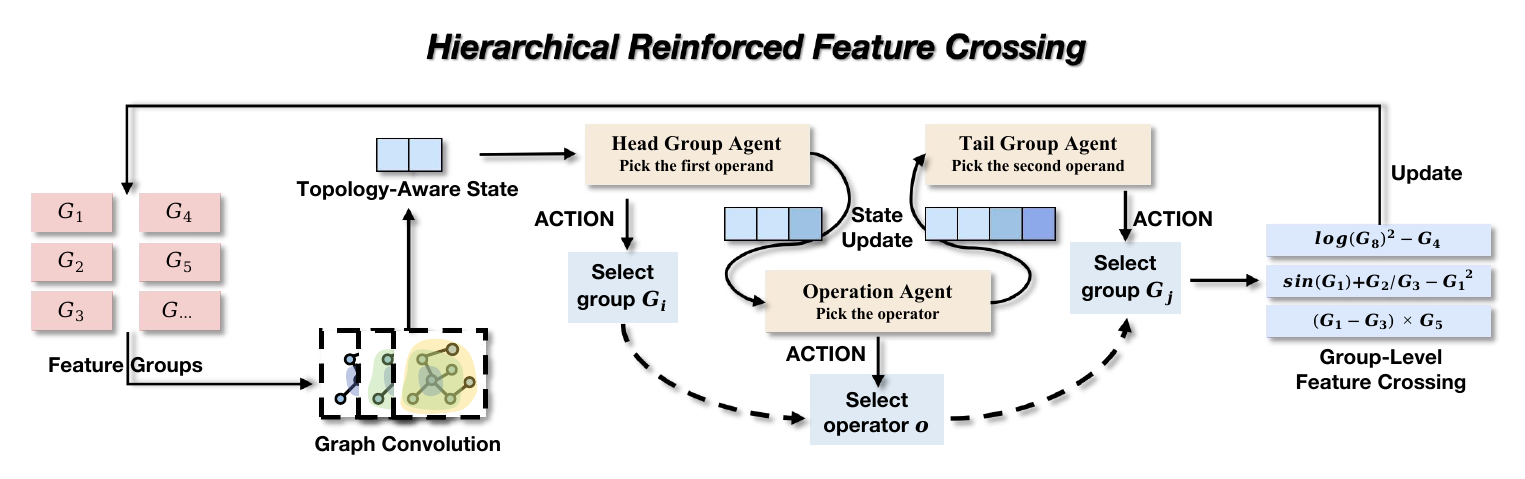}
    \caption{Hierarchical reinforced feature crossing. We develop three reinforcement agents to select feature groups and an operation, and then cross the feature groups to generate new features in an iterative way.}
    \label{hierarchy}
\end{figure}
\textbf{Why hierarchical reinforcement feature crossing matters?} Firstly, explicit feature crosses can generate more interpretable and meaningful new dimensions to form a new feature space. For instance, in a movie recommendation system, crossing features describing age and gender produces a meaningful new dimension to describe subpopulations of users interested in certain types of movies (e.g., 20-year-old females might prefer watching romantic dramas). 
Secondly, the interaction gain between features is crucial in determining whether to cross features; however, measuring such implicit information gain is challenging. Therefore, reinforcement learning serves as a valuable tool to perform AI based decision-making when feature space mechanisms are unclear. 
Moreover, feature crossing is based on previously generated features and changes the state of a feature space over time, thus we can formulate it into a Markov decision process. 
We construct three agents: two for selecting features and one for selecting operations. A simple design is all the three agents are independent. However, we find that feature agents and operation agents mutually influence each other in a cascading fashion, forming a hierarchical structure.  Therefore, we use hierarchical reinforcement learning to cross new features.

\noindent\textbf{Leveraging self-optimizing reinforcement and hierarchical agent structures to cross features.} 
We propose to develop a hierarchical reinforcement crossing framework that includes two feature controllers and one operation controller, each of which in each iteration takes a state as input, conducts an action, and generates a new feature to obtain an updated feature set, as shown in Figure~\ref{hierarchy}. For example, in each iteration, a feature controller firstly selects a feature from the current feature set. Secondly, the operation controller selects an operation based on both the current feature set and the chosen feature. Thirdly, another feature controller selects a feature based on the previous selections. By iterating the three steps, we generate an improved feature set and feed it into a downstream ML model to obtain predictive performances as a reward. The key components of our proposed method are as follows:

\noindent\emph{1) State.} We input the fixed-length embedding mentioned in Section~\ref{GNN-state} into the RL system as a state, which enables us to capture the crucial topological information of graphs. 

\noindent\emph{2) Agents.} We develop a hierarchical structure with three different agents to select features and then generate new features in each iteration, and each agent is implemented by a deep Q-network (DQN)~\cite{DQN}. Here we take the $t$-th iteration as an example:
i) \textbf{Head feature agent}. Given a subgraph with a node feature matrix $\mathbf{X}^{t-1} = [G_1, G_2, ..., G_k]$ generated by the last iteration, where $G$ represents a feature group. We use a pre-trained GNN to transfer $\mathbf{X}^{t-1}$ to a vector that represents a state, denoted as $g(\mathbf{X}^{t-1})$. The action is to select a head feature group $G_i$ from $\mathbf{X}^{t-1}$ by the reinforced agent. 
ii) \textbf{Operation agent}. The state includes the representation of $\mathbf{X}^{t-1}$ and the head feature group $G_i$, denoted as $g(\mathbf{X}^{t-1}) \oplus g(G_i)$. The action is to select an operation $o$ from the operation set by this reinforced agent. 
iii) \textbf{Tail feature agent}. The state includes the representation of $\mathbf{X}^{t-1}$, the head feature group $G_i$ and the selected operation $o$, denoted as $g(\mathbf{X}^{t-1}) \oplus g(G_i) \oplus rep(o)$, where $rep(\cdot)$ is a function to convert an operation into a vector. The action is to select a tail feature group $G_j$ from $\mathbf{X}^{t-1}$ by this agent. 

\noindent\emph{3) Rewards.} When the three agents select two feature groups and an operation: i) If the operator is unary, we directly use $G_i$ and $o$ for feature crossing; ii) If the operator is binary, we select features from $G_i$ and $G_j$ to form feature pairs and then cross them using $o$. We choose the top k features based on the relevance between the crossed features and target labels to add to the original feature set, forming a new feature set $\mathbf{X}^t$.
We calculate the reward based on the performance of downstream ML tasks, aiming to evaluate whether the current selections of agents contribute to the information gain of the generated feature set.  
For a good feature set, each feature should contain as much unique information as possible, resulting in an improved performance of the supervised downstream task. Therefore, for a generated feature set in the current iteration, the reward is defined as:
\begin{equation}
    r(\mathbf{X}^t, y) = \mathcal{V}_{\mathcal{M}}(\mathcal{A}(\mathbf{X}^t), y) - p, 
\end{equation}
where $\mathcal{V}_{\mathcal{M}}$ is the performance (e.g., F1-score) of a downstream task, $\mathcal{A}$ means apply the cross to the whole graph, and $p$ is the best performance updated with the iteration. We assign the reward to these three agents: i) Head feature agent: $r_i = w_1*r(\mathbf{X}^t, y)$; ii) Operation agent: $r_o = w_2*r(\mathbf{X}^t, y)$; iii) Tail feature agent: $r_j = w_3*r(\mathbf{X}^t, y)$, where $w_i(i \in 1,2,3)$ is a positive weight.

\noindent\emph{4) Policy.} We train the three agents by maximizing the discounted and cumulative reward during the iterative feature generation process. We encourage the hierarchical agents to collaborate to generate a feature set that is informative and performs well in the downstream ML task. To simplify equations, we use general notations to illustrate the optimization process of any agent. Here, $s_t$, $a_t$, and $r_t$ represent the current state, action, and reward respectively. We minimize the temporal difference error $\mathcal{L}$ converted from the Bellman equation~\cite{bellman}, given by:
\begin{equation}
    \mathcal{L} = Q(s_t, a_t) - (r_t + \gamma * \max_{a_{t+1}}Q(s_{t+1}, a_{t+1})),
\end{equation}
where $\gamma \in [0,1]$ is the discounted factor, and $Q$ is the $Q$ function estimated by the DQN. Upon the convergence of the agent, our expectation is to discover the optimal policy $\pi^*$, which selects the best action according to the state through the Q-value, which can be formulated as follows:
\begin{equation}
    \pi^* = \arg\max_a(s_t, a).
\end{equation}
\section{Experimental Results}

In this section, we evaluate our method with baselines on node-level, edge-level, and graph-level tasks of graphs. In particular, we wish to answer the following research questions: 
\textbf{Q1:} How effective is our method under the four datasets for three graph tasks?
\textbf{Q2:} How does the subgraphs mining influence both effectiveness and efficiency?
\textbf{Q3:} How does the GNN-based state influence the performance?
\textbf{Q4:} How does feature grouping affect the outcomes?
\textbf{Q5:} What is the comparative training time complexity of our method versus the baselines?
\textbf{Q6:} What the reconstructed feature space looks like?

\setlength{\tabcolsep}{0.7mm}{
\begin{table}[h]
\centering
\small
\caption{Statistics of the datasets.}
\begin{tabular}{@{}cccccccccc@{}}
\toprule\toprule
Datasets       & \#Graphs & \#Nodes & Avg. \#Nodes & Avg. \#Edges & \#Attributes & \#Graph Classes & \#Node Classes & \#Graph Labels & \#Node Labels  \\ \midrule
ENZYMES        & 600            & 19580          & 32.63    & 62.14    & 18   & 6          & 3   & \CheckmarkBold &  \CheckmarkBold  \\
PROTEINS & 1113           & 43471          & 39.06    & 72.82    & 29   & 2          & 3         & \CheckmarkBold & \CheckmarkBold \\
Synthie        & 400            & 38000          & 95.00    & 172.93   & 15   & 4          & -   & \CheckmarkBold & \XSolidBrush    \\
AIDS           & 2000           & 31385          & 15.69    & 16.20    & 4    & 2          & 38  & \CheckmarkBold & \CheckmarkBold      \\
\bottomrule\bottomrule
\end{tabular}
\label{data_statistic}
\end{table}}

\subsection{Experimental Settings}
\textbf{Statistics of the datasets.} We use 4 public benchmark datasets to conduct experiments. These datasets are from different data domains:
1) Bioinformatics dataset: ENZYMES~\cite{enzyme}, PROTEINS~\cite{proteins};
2) Synthetic dataset: Synthie~\cite{synthie};
3) Small molecules dataset: AIDS~\cite{aids}. These datasets are publicly available at \footnote{https://ls11-www.cs.tu-dortmund.de/staff/morris/graphkerneldatasets}. 
Statistics of the datasets are summarized in Table~\ref{data_statistic}. We compute the total number of graphs, the number of nodes, the number of average nodes, the number of average edges, the class of graphs, and the class of nodes for each dataset. 

\noindent\textbf{Evaluation.}
We test the transformed graphs with three widely used graph downstream tasks: 
1) Node Classification;
2) Link Prediction;
3) Graph Classification.
To ensure fair comparisons and alleviate downstream ML model variance, we utilize a simple Multi-Layer Perceptron (MLP) with fixed hyperparameters and a consistent seed for each task to ensure reproducibility. For link prediction, we randomly sample negative edges with the same number as positive ones. Each dataset is split into 80\% training and 20\% testing subsets. F1-score, Precision, and Recall are employed as evaluation metrics for the three tasks, and Area Under Curve (AUC) is also used for link prediction. Higher metric values denote higher graph feature space quality. To ensure consistency, we randomly executed our method 10 times, and overall performance is reported as mean plus standard deviation.

\noindent\textbf{Reproducibility.} 1) Core Frequent Subgraph Mining: We retain the subgraphs if they appear in the original graphs in more than 10\% of all graphs and the subgraphs exit 3 nodes at least.
2) Reinforcement Feature Space Reconstruction: We limited iterations (epochs) to 10, with each iteration consisting of 10 exploration steps. When the number of generated features is fourth of the original feature set size, we performed a feature selection to control the feature set size. All agents were conducted using a DQN with two linear layers activated by ReLU function. We use the Adam optimizer with a 0.01 learning rate, and set the limit of the experience replay memory as 32 and the batch size as 8. The parameters of all baseline models are set based on the default settings of corresponding papers. More details are released in the code.

\noindent\textbf{Environmental Settings.} All experiments are conducted on the Ubuntu 22.04.3 LTS operating system, Intel(R) Core(TM) i9-13900KF CPU@ 3GHz, and 1 way RTX 4090 and 32GB of RAM, with the framework of Python 3.11.4 and PyTorch 2.0.1.

\noindent\textbf{Baseline Algorithms.}
We compare our method with widely used feature transformation methods:
1) \textbf{RDG} randomly generates feature-operation-feature transformation records to create a new feature space;
2) \textbf{PCA}~\cite{uddin2021pca} uses linear feature correlation to generate new features;
3) \textbf{LDA}~\cite{LDA} is a matrix factorization-based method that obtains decomposed latent representation as the generated feature space; 
4) \textbf{ERG} first applies a set of operations to each feature to expand the feature space, and then selects key features from it to form a new feature space; 
5) \textbf{NFS}~\cite{NFS} embeds the decision-making process of feature transformation into a policy network and utilizes reinforcement learning (RL) to optimize the entire feature transformation process; 
6) \textbf{TTG}~\cite{TTG} represents the feature transformation process as a graph and implements an RL-based discrete search method to find the optimal feature set.

\subsection{Overall Performance (RQ1)}
\setlength{\tabcolsep}{1.2mm}{
\begin{table*}[tb]
\centering
\fontsize{6}{9}\selectfont
\caption{Overall Performance. Mean values with standard deviation by 10 repeats with different seeds. A higher value denotes a higher reconstructed feature space quality.}
\begin{tabular}{@{}c|c|ccc|cccc|ccc@{}}
\toprule
\multirow{2}{*}{\rotatebox{90}{Datasets}} & \multirow{2}{*}{Methods} & \multicolumn{3}{c|}{Node Classification}                                       & \multicolumn{4}{c|}{Link Prediction}                                                                                            & \multicolumn{3}{c}{Graph Classfication}                                          \\ \cmidrule(l){3-12} 
                          &                          & Precision                & Recall                   & F1                       & Precision                           & Recall                   & F1                                  & AUC                      & Precision                & Recall                     & F1                       \\ \midrule
\multirow{8}{*}{\rotatebox{90}{ENZYMES}}  & Origin                   & 0.899$\pm$0.000          & 0.905$\pm$0.000          & 0.895$\pm$0.000          & 0.644$\pm$0.000                     & 0.644$\pm$0.000          & 0.643$\pm$0.000                     & 0.644$\pm$0.000          & 0.193$\pm$0.000          & 0.292$\pm$0.000            & 0.229$\pm$0.000          \\
                          & PCA                      & 0.753$\pm$0.000          & 0.768$\pm$0.000          & 0.756$\pm$0.000          & 0.581$\pm$0.004                     & 0.581$\pm$0.004          & 0.579$\pm$0.004                     & 0.580$\pm$0.004          & 0.239$\pm$0.000          & 0.292$\pm$0.000            & 0.260$\pm$0.000          \\
                          & LDA                      & 0.682$\pm$0.008          & 0.695$\pm$0.006          & 0.684$\pm$0.007          & 0.565$\pm$0.008                     & 0.564$\pm$0.008          & 0.563$\pm$0.008                     & 0.564$\pm$0.008          & 0.217$\pm$0.048          & 0.240$\pm$0.025            & 0.212$\pm$0.027          \\
                          & RDG                      & 0.917$\pm$0.004          & 0.926$\pm$0.003          & 0.918$\pm$0.003          & 0.651$\pm$0.005                     & 0.646$\pm$0.004          & 0.643$\pm$0.005                     & 0.646$\pm$0.004          & 0.265$\pm$0.014          & 0.308$\pm$0.10             & 0.278$\pm$0.016          \\
                          & ERG                      & 0.923$\pm$0.004          & 0.930$\pm$0.003          & 0.923$\pm$0.003          & 0.651$\pm$0.005                     & 0.646$\pm$0.005          & 0.643$\pm$0.005                     & 0.646$\pm$0.005          & 0.271$\pm$0.021          & 0.320$\pm$0.016            & 0.285$\pm$0.020          \\
                          & NFS                      & 0.922$\pm$0.003          & 0.930$\pm$0.002          & 0.922$\pm$0.002          & 0.658$\pm$0.006                     & 0.653$\pm$0.005          & 0.651$\pm$0.005                     & 0.653$\pm$0.005          & 0.208$\pm$0.000          & 0.317$\pm$0.000            & 0.250$\pm$0.000          \\
                          & TTG                      & 0.919$\pm$0.003          & 0.928$\pm$0.002          & 0.920$\pm$0.003          & 0.657$\pm$0.008                     & 0.654$\pm$0.005          & 0.647$\pm$0.005                     & 0.654$\pm$0.004          & 0.257$\pm$0.042          & 0.308$\pm$0.015            & 0.261$\pm$0.026          \\
                          & \textbf{\model}          & \textbf{0.936$\pm$0.006} & \textbf{0.943$\pm$0.005} & \textbf{0.937$\pm$0.006} & \textbf{0.662$\pm$0.005}            & \textbf{0.660$\pm$0.005} & \textbf{0.660$\pm$0.005}            & \textbf{0.660$\pm$0.005} & \textbf{0.342$\pm$0.053} & \textbf{0.358$\pm$0.029}   & \textbf{0.325$\pm$0.029} \\ \midrule
\multirow{8}{*}{\rotatebox{90}{PROTEINS}} & Origin                   & 0.790$\pm$0.000          & 0.815$\pm$0.000          & 0.802$\pm$0.000          & 0.982$\pm$0.000                     & 0.981$\pm$0.000          & 0.981$\pm$0.000                     & 0.981$\pm$0.000          & 0.746$\pm$0.000          & 0.748$\pm$0.000            & 0.747$\pm$0.000          \\
                          & PCA                      & 0.853$\pm$0.000          & 0.849$\pm$0.000          & 0.833$\pm$0.000          & 0.952$\pm$0.002                     & 0.947$\pm$0.002          & 0.947$\pm$0.002                     & 0.947$\pm$0.002          & 0.654$\pm$0.000          & 0.658$\pm$0.000            & 0.655$\pm$0.000          \\
                          & LDA                      & 0.708$\pm$0.048          & 0.719$\pm$0.048          & 0.706$\pm$0.049          & 0.652$\pm$0.012                     & 0.647$\pm$0.015          & 0.644$\pm$0.018                     & 0.647$\pm$0.015          & 0.620$\pm$0.023          & 0.630$\pm$0.020            & 0.614$\pm$0.022          \\
                          & RDG                      & 0.882$\pm$0.013          & 0.909$\pm$0.013          & 0.895$\pm$0.013          & 0.984$\pm$0.002                     & 0.984$\pm$0.002          & 0.984$\pm$0.002                     & 0.984$\pm$0.002          & 0.751$\pm$0.003          & 0.750$\pm$0.002            & 0.749$\pm$0.005          \\
                          & ERG                      & 0.834$\pm$0.020          & 0.845$\pm$0.023          & 0.835$\pm$0.020          & 0.981$\pm$0.003                     & 0.980$\pm$0.003          & 0.980$\pm$0.003                     & 0.980$\pm$0.003          & 0.758$\pm$0.013          & 0.756$\pm$0.011            & 0.753$\pm$0.010          \\
                          & NFS                      & 0.821$\pm$0.001          & 0.830$\pm$0.005          & 0.821$\pm$0.006          & 0.983$\pm$0.001                     & 0.983$\pm$0.001          & 0.982$\pm$0.001                     & 0.983$\pm$0.001          & 0.751$\pm$0.004          & 0.751$\pm$0.000            & 0.751$\pm$0.000          \\
                          & TTG                      & 0.847$\pm$0.005          & 0.856$\pm$0.005          & 0.847$\pm$0.005          & 0.984$\pm$0.002                     & 0.984$\pm$0.002          & 0.984$\pm$0.002                     & 0.984$\pm$0.002          & 0.750$\pm$0.004          & 0.751$\pm$0.005            & 0.751$\pm$0.004          \\
                          & \textbf{\model}          & \textbf{0.916$\pm$0.017} & \textbf{0.925$\pm$0.017} & \textbf{0.916$\pm$0.017} & \textbf{0.988$\pm$0.002}            & \textbf{0.988$\pm$0.002} & \textbf{0.988$\pm$0.001}            & \textbf{0.988$\pm$0.001} & \textbf{0.767$\pm$0.006} & \textbf{0.766$\pm$0.0.007} & \textbf{0.765$\pm$0.006} \\ \midrule
\multirow{8}{*}{\rotatebox{90}{Synthie}}  & Origin                   & -                        & -                        & -                        & 0.499$\pm$0.000                     & 0.499$\pm$0.000          & 0.499$\pm$0.000                     & 0.500$\pm$0.000          & 0.490$\pm$0.000          & 0.487$\pm$0.000            & 0.487$\pm$0.000          \\
                          & PCA                      & -                        & -                        & -                        & 0.508$\pm$0.003                     & 0.508$\pm$0.003          & 0.507$\pm$0.004                     & 0.508$\pm$0.003          & 0.461$\pm$0.000          & 0.462$\pm$0.000            & 0.436$\pm$0.000          \\
                          & LDA                      & -                        & -                        & -                        & 0.504$\pm$0.008                     & 0.504$\pm$0.008          & 0.499$\pm$0.011                     & 0.504$\pm$0.008          & 0.128$\pm$0.113          & 0.221$\pm$0.057            & 0.121$\pm$0.083          \\
                          & RDG                      & -                        & -                        & -                        & 0.515$\pm$0.004                     & 0.515$\pm$0.004          & 0.514$\pm$0.004                     & 0.515$\pm$0.004          & 0.484$\pm$0.010          & 0.470$\pm$0.010            & 0.465$\pm$0.019          \\
                          & ERG                      & -                        & -                        & -                        & 0.514$\pm$0.004                     & 0.513$\pm$0.004          & 0.513$\pm$0.004                     & 0.513$\pm$0.004          & 0.496$\pm$0.027          & 0.477$\pm$0.022            & 0.473$\pm$0.025          \\
                          & NFS                      & -                        & -                        & -                        & 0.510$\pm$0.006                     & 0.510$\pm$0.006          & 0.510$\pm$0.006                     & 0.510$\pm$0.006          & 0.540$\pm$0.091          & 0.465$\pm$0.029            & 0.412$\pm$0.029          \\
                          & TTG                      & -                        & -                        & -                        & 0.511$\pm$0.004                     & 0.511$\pm$0.004          & 0.510$\pm$0.004                     & 0.511$\pm$0.004          & 0.501$\pm$0.021          & 0.493$\pm$0.018            & 0.493$\pm$0.018          \\
                          & \textbf{\model}          & -                        & -                        & -                        & \textbf{0.523$\pm$0.002}            & \textbf{0.522$\pm$0.002} & \textbf{0.522$\pm$0.002}            & \textbf{0.521$\pm$0.003} & \textbf{0.571$\pm$0.032} & \textbf{0.556$\pm$0.026}   & \textbf{0.551$\pm$0.026} \\ \midrule
\multirow{8}{*}{\rotatebox{90}{AIDS}}     & Origin                   & 0.892$\pm$0.000          & 0.929$\pm$0.000          & 0.906$\pm$0.000          & 0.922$\pm$0.000                     & 0.913$\pm$0.000          & 0.913$\pm$0.000                     & 0.913$\pm$0.000          & 0.871$\pm$0.000          & 0.875$\pm$0.000            & 0.864$\pm$0.000          \\
                          & PCA                      & 0.381$\pm$0.000          & 0.615$\pm$0.000          & 0.471$\pm$0.000          & 0.821$\pm$0.017                     & 0.790$\pm$0.011          & 0.784$\pm$0.010                     & 0.790$\pm$0.010          & 0.899$\pm$0.000          & 0.899$\pm$0.000            & 0.893$\pm$0.000          \\
                          & LDA                      & 0.595$\pm$0.083          & 0.741$\pm$0.027          & 0.649$\pm$0.049          & 0.727$\pm$0.017                     & 0.682$\pm$0.013          & 0.666$\pm$0.019                     & 0.682$\pm$0.013          & 0.887$\pm$0.005          & 0.889$\pm$0.004            & 0.882$\pm$0.006          \\
                          & RDG                      & 0.957$\pm$0.029          & 0.969$\pm$0.019          & 0.961$\pm$0.025          & 0.921$\pm$0.003                     & 0.915$\pm$0.004          & 0.915$\pm$0.004                     & 0.916$\pm$0.004          & 0.938$\pm$0.023          & 0.938$\pm$0.022            & 0.937$\pm$0.023          \\
                          & ERG                      & 0.953$\pm$0.029          & 0.966$\pm$0.019          & 0.958$\pm$0.025          & 0.920$\pm$0.003                     & 0.914$\pm$0.003          & 0.915$\pm$0.004                     & 0.915$\pm$0.004          & 0.937$\pm$0.021          & 0.938$\pm$0.020            & 0.936$\pm$0.021          \\
                          & NFS                      & 0.906$\pm$0.002          & 0.937$\pm$0.001          & 0.918$\pm$0.001          & 0.924$\pm$0.003                     & 0.916$\pm$0.004          & 0.916$\pm$0.004                     & 0.916$\pm$0.004          & 0.900$\pm$0.005          & 0.902$\pm$0.005            & 0.898$\pm$0.005          \\
                          & TTG                      & 0.925$\pm$0.023          & 0.947$\pm$0.016          & 0.933$\pm$0.020          & 0.925$\pm$0.002                     & 0.917$\pm$0.003          & 0.917$\pm$0.003                     & 0.918$\pm$0.003          & 0.896$\pm$0.000          & 0.899$\pm$0.000            & 0.893$\pm$0.000          \\
                          & \textbf{\model}          & \textbf{0.988$\pm$0.002} & \textbf{0.991$\pm$0.001} & \textbf{0.988$\pm$0.002} & \textbf{0.933$\pm$0.002}            & \textbf{0.927$\pm$0.003} & \textbf{0.927$\pm$0.003}            & \textbf{0.927$\pm$0.003} & \textbf{0.984$\pm$0.002} & \textbf{0.984$\pm$0.002}   & \textbf{0.984$\pm$0.002} \\ \bottomrule
\end{tabular}
\label{exp:overall_performance}
\end{table*}}
In this experiment, we aim to respond to research question 1: can our method effectively reconstruct quality feature space to improve downstream tasks? We compare the overall performance of {\model} in terms of Precision, Recall, F1-score, and AUC in three wildly used graph tasks. Table~\ref{exp:overall_performance} shows {\model} performs the best in all datasets across different downstream tasks. The underlying driver for this observation is that {\model} can accurately capture intricate and valuable topological information through core subgraph mining and GNN-based embedding, and then the hierarchical graph feature crossing policies in {\model} incorporate the graph information, feature information distinctness, and valuable feedback to generate informative features for a graph. In conclusion, this experiment validates that our method is more effective in graph feature space reconstruction.

\setlength{\tabcolsep}{0.7mm}{
\begin{table}[]
\centering
\fontsize{6}{9}\selectfont
\caption{The impact of subgraph. Mean values with standard deviation by 10 repeats.}
\begin{tabular}{@{}c|c|ccccc|ccccc@{}}
\toprule\toprule
Dataset                    & Downstream Task      & \multicolumn{5}{c|}{\model}                                                           & \multicolumn{5}{c}{\model$^-$}                                        \\ \midrule
                           &                      & Precision       & Recall            & F1              & AUC             & Time Cost (s)  & Precision       & Recall          & F1              & AUC             & Time Cost (s)  \\ \midrule
\multirow{3}{*}{ENZYMES}   & Node Classification  & 0.936$\pm$0.006 & 0.943$\pm$0.005   & 0.937$\pm$0.006 & -               & 279$\pm$62 & 0.936$\pm$0.004 & 0.941$\pm$0.004 & 0.934$\pm$0.005 & -               & 413$\pm$84  \\
                           & Link Prediction      & 0.662$\pm$0.005 & 0.660$\pm$0.005   & 0.660$\pm$0.005 & 0.660$\pm$0.005 & 271$\pm$66 & 0.655$\pm$0.005 & 0.655$\pm$0.005 & 0.655$\pm$0.005 & 0.655$\pm$0.005 & 503$\pm$109 \\
                           & Graph Classification & 0.342$\pm$0.053 & 0.358$\pm$0.029   & 0.325$\pm$0.029 & -               & 148$\pm$26 & 0.306$\pm$0.023 & 0.338$\pm$0.028 & 0.309$\pm$0.028 & -               & 277$\pm$52  \\ \midrule
\multirow{3}{*}{PRROTEINS} & Node Classification  & 0.916$\pm$0.017 & 0.925$\pm$0.017   & 0.916$\pm$0.017 & -               & 385$\pm$47 & 0.896$\pm$0.042 & 0.906$\pm$0.044 & 0.895$\pm$0.043 & -               & 883$\pm$213 \\
                           & Link Prediction      & 0.988$\pm$0.002 & 0.988$\pm$0.002   & 0.988$\pm$0.001 & 0.988$\pm$0.001 & 313$\pm$22 & 0.985$\pm$0.002 & 0.985$\pm$0.002 & 0.985$\pm$0.002 & 0.985$\pm$0.002 & 357$\pm$68  \\
                           & Graph Classification & 0.767$\pm$0.006 & 0.766$\pm$0.007   & 0.765$\pm$0.006 & -               & 258$\pm$28 & 0.759$\pm$0.004 & 0.759$\pm$0.005 & 0.759$\pm$0.005 & -               & 329$\pm$44  \\ \midrule
\multirow{3}{*}{Synthie}   & Node Classification  & -               & -                 & -               & -               & -          & -               & -               & -               & -               & -           \\
                           & Link Prediction      & 0.523$\pm$0.002 & 0.522$\pm$0.002   & 0.522$\pm$0.002 & 0.522$\pm$0.002 & 397$\pm$93 & 0.514$\pm$0.005 & 0.514$\pm$0.005 & 0.513$\pm$0.005 & 0.514$\pm$0.005 & 505$\pm$30  \\
                           & Graph Classification & 0.521$\pm$0.042 & 0.571$\pm$0.032   & 0.556$\pm$0.026 & -               & 278$\pm$40 & 0.517$\pm$0.017 & 0.518$\pm$0.017 & 0.513$\pm$0.011 & -               & 363$\pm$50  \\ \midrule
\multirow{3}{*}{AIDS}      & Node Classification  & 0.988$\pm$0.002 & 0.991$\pm$0.001   & 0.988$\pm$0.002 & -               & 264$\pm$77 & 0.981$\pm$0.011 & 0.985$\pm$0.009 & 0.982$\pm$0.010 & -               & 282$\pm$69  \\
                           & Link Prediction      & 0.933$\pm$0.002 & 0.927$\pm$0.003   & 0.927$\pm$0.003 & 0.927$\pm$0.003 & 165$\pm$61 & 0.930$\pm$0.002 & 0.923$\pm$0.003 & 0.923$\pm$0.003 & 0.923$\pm$0.003 & 210$\pm$47  \\
                           & Graph Classification & 0.984$\pm$0.002 & 0.984$\pm$0.002   & 0.984$\pm$0.002 & -               & 249$\pm$49 & 0.980$\pm$0.002 & 0.980$\pm$0.002 & 0.980$\pm$0.002 & -               & 333$\pm$55  \\ \bottomrule\bottomrule
\end{tabular}
\label{table:subgraph}
\end{table}}
\begin{figure}[]
    \centering
    \subfigure[Node Cls]{\includegraphics[width=0.33\textwidth]{{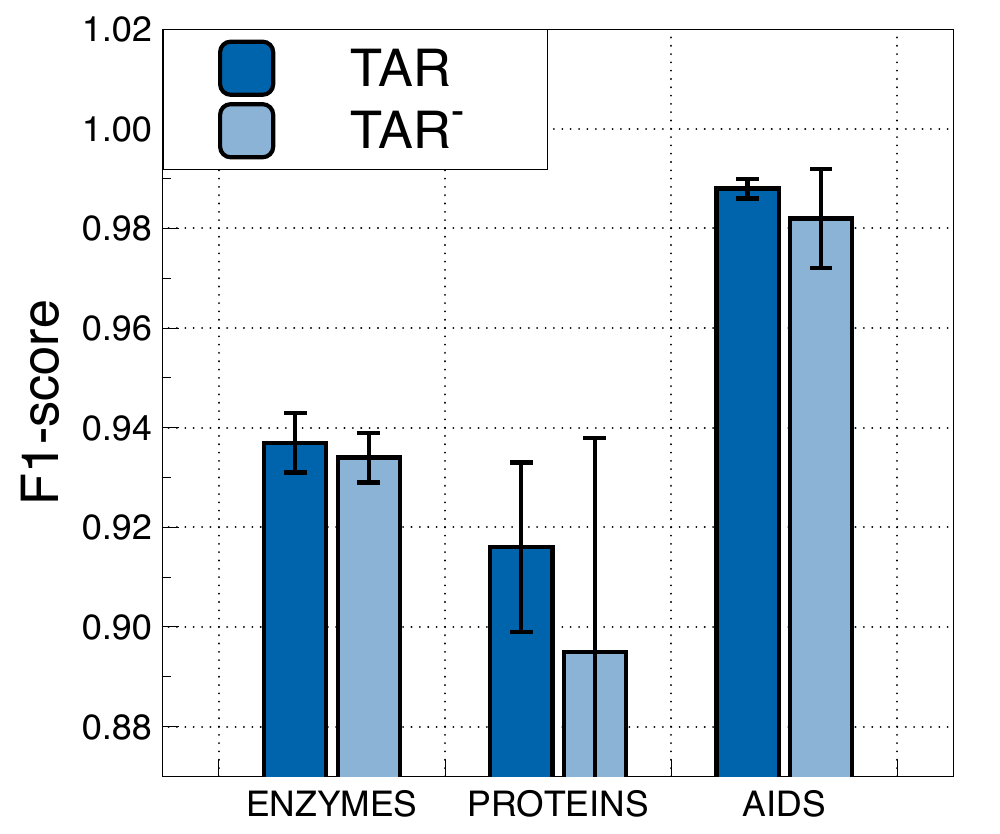}}}
    \subfigure[Link Pre]{\includegraphics[width=0.33\textwidth]{{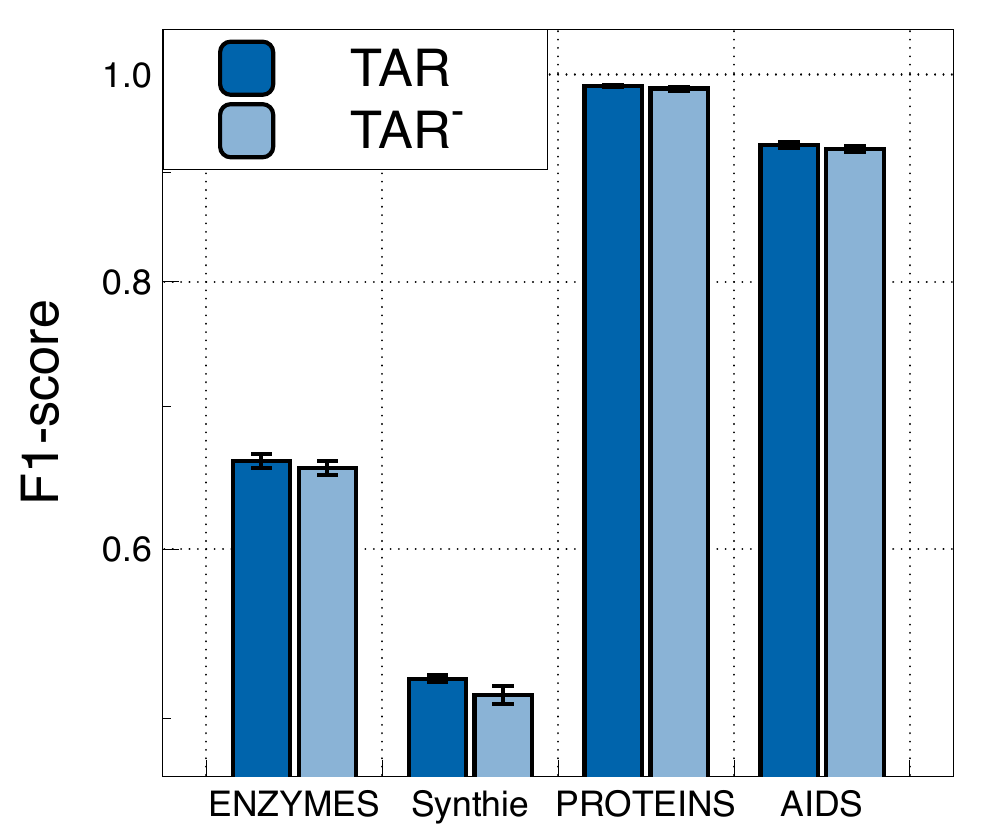}}}
    \subfigure[Graph Cls]{\includegraphics[width=0.33\textwidth]{{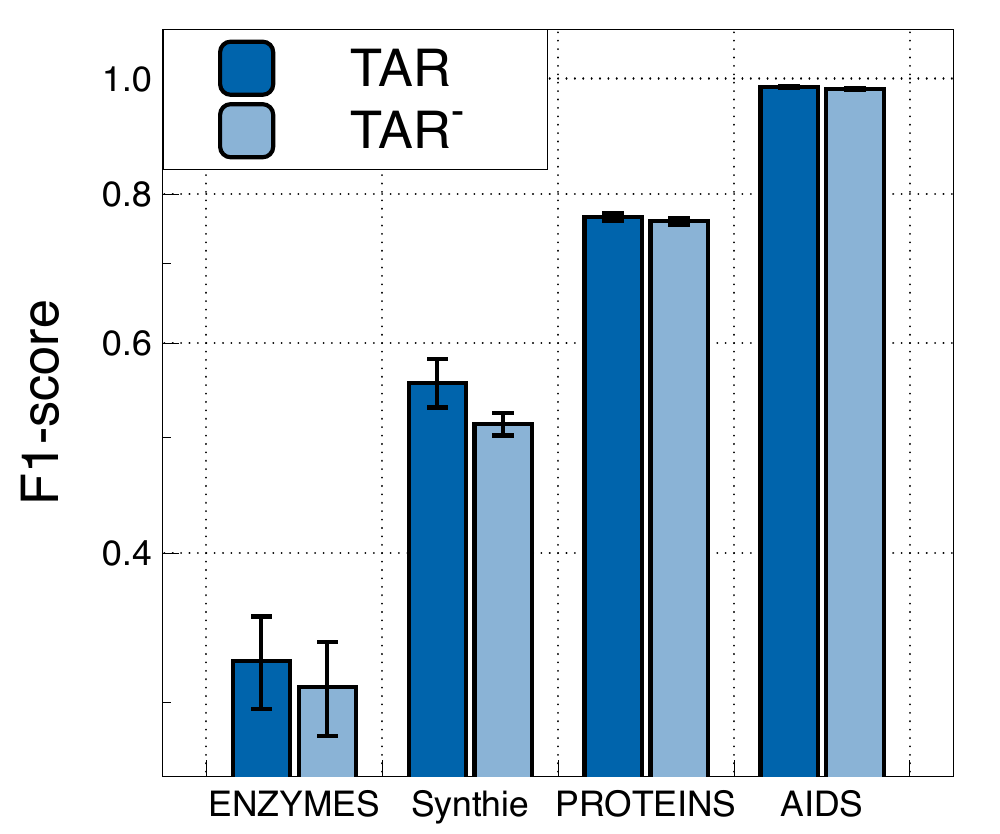}}}
    \subfigure[Node Cls Time]{\includegraphics[width=0.33\textwidth]{{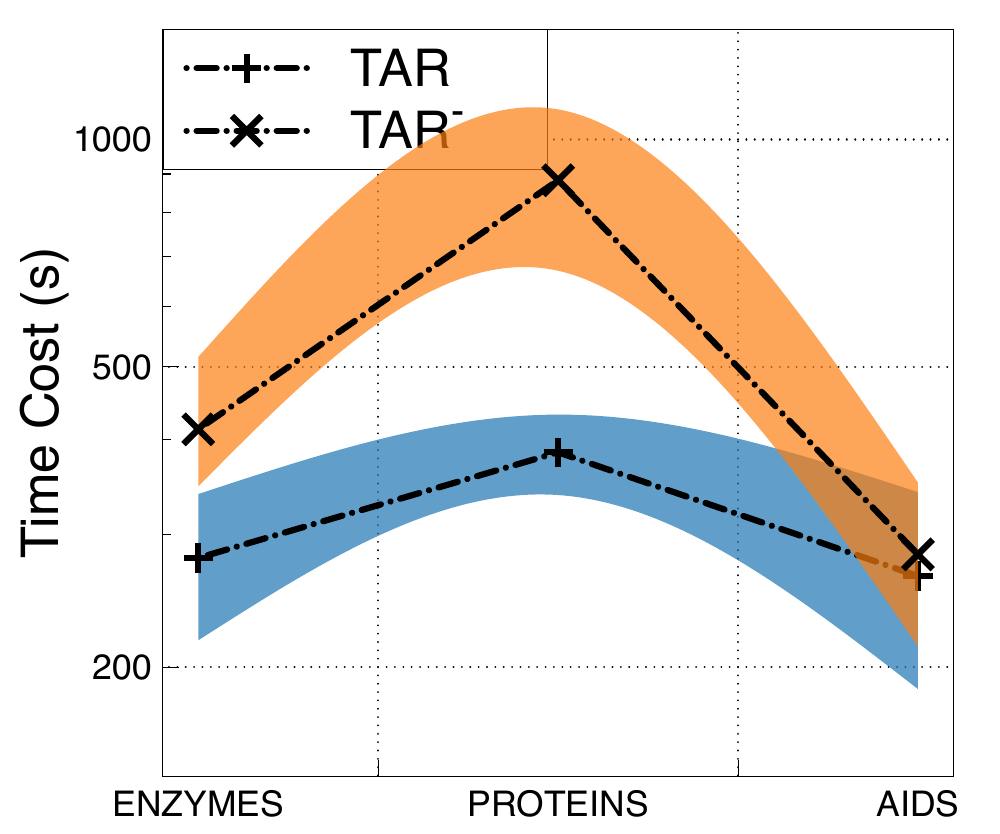}}}
    \subfigure[Link Pre Time]{\includegraphics[width=0.33\textwidth]{{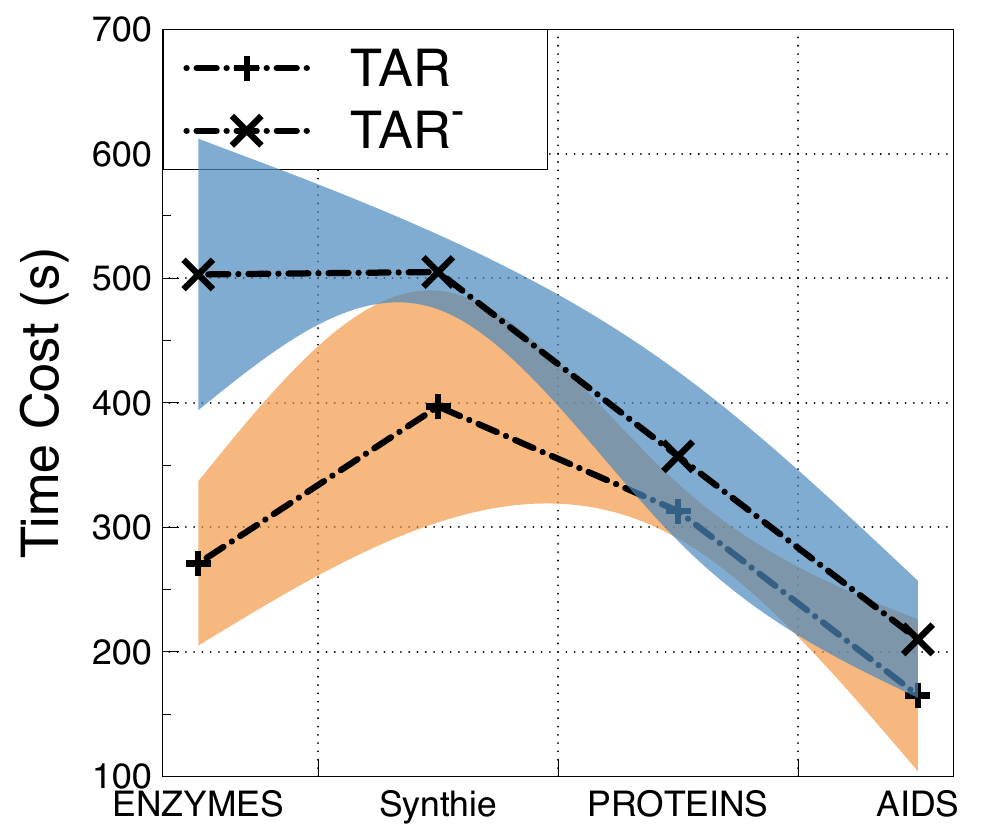}}}
    \subfigure[Graph Cls Time]{\includegraphics[width=0.33\textwidth]{{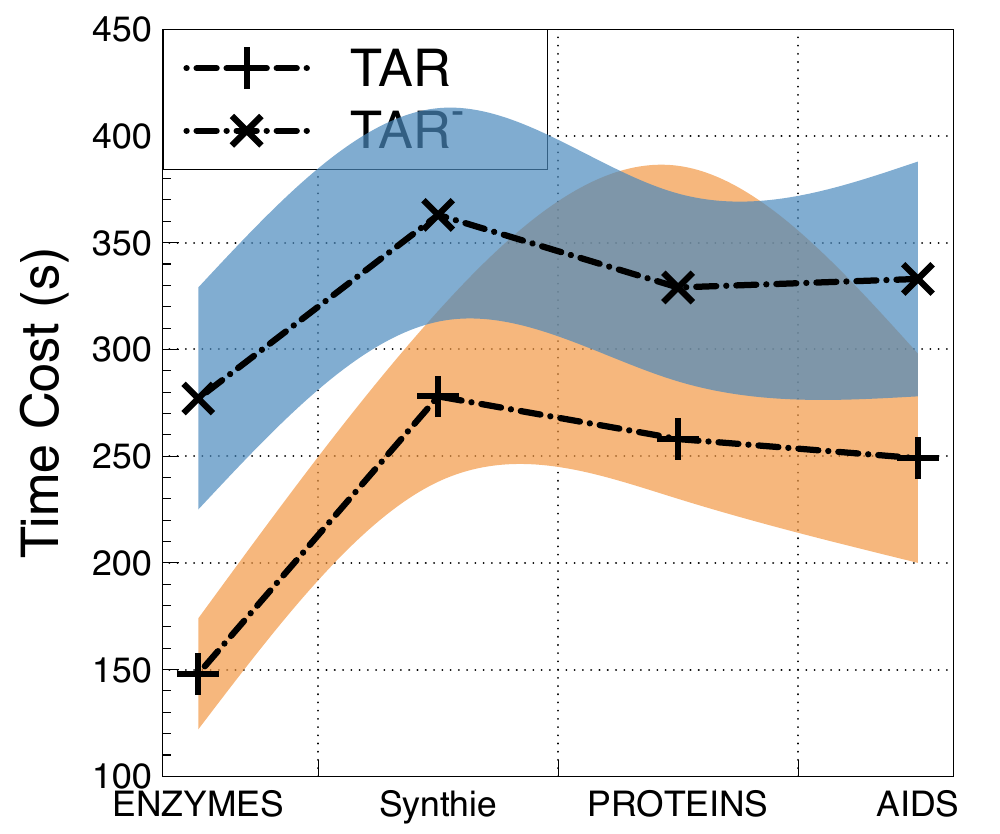}}}
    \caption{The impact of subgraph. Subfigure (a), (b), and (c) are F1-score comparisons of three different tasks across all datasets. Subfigure (d), (e), and (f) are training time cost comparisons of three different tasks across all datasets. We report mean values with standard deviation by 10 repeats.}
    \label{impact-subgraph}
\end{figure}
\subsection{Impact of the Subgraph (RQ2)}
One of the most important novelties of {\model} is to involve subgraph mining to focus on the most informative and critical topological structures. To analyze the effectiveness and efficiency of this component, we develop a variant model {\model$^-$}, which uses the complete graph data instead of subgraphs to reconstruct the feature space. Figure~\ref{impact-subgraph} (a)(b)(c) show that {\model} can achieve better performance compared to {\model$^-$}. The underlying driver is that {\model} crosses features based on subgraphs, which denoises the graph and enables the crossing strategy tailored to the most critical structures and features, thus better capturing the intrinsic characteristics. Another observation is that {\model} achieves better efficiency in terms of training time cost, as shown in Figure~\ref{impact-subgraph} (d)(e)(f). The reason is that transforming a complex graph into a simplified subgraph eases the model computation, which reduces the computational complexity. In summary, this experiment demonstrates the effectiveness and efficiency of including subgraphs. The detailed experiment results for all datasets are reported in Table~\ref{table:subgraph}.

\setlength{\tabcolsep}{2.1mm}{
\begin{table}[h]
\centering
\fontsize{6}{9}\selectfont
\caption{The impact of GNN-based state. Mean values with standard deviation by 10 repeats.}
\begin{tabular}{@{}c|c|cccc|cccc@{}}
\toprule\toprule
Dataset                    & Downstream Task      & \multicolumn{4}{c|}{\model}                                              & \multicolumn{4}{c}{\model$^+$}                          \\ \midrule
                           &                      & Precision       & Recall            & F1              & AUC             & Precision       & Recall          & F1              & AUC             \\ \midrule
\multirow{3}{*}{ENZYMES}   & Node Classification  & 0.936$\pm$0.006 & 0.943$\pm$0.005   & 0.937$\pm$0.006 & -               & 0.928$\pm$0.007 & 0.935$\pm$0.006 & 0.929$\pm$0.007 & -               \\
                           & Link Prediction      & 0.662$\pm$0.005 & 0.660$\pm$0.005   & 0.660$\pm$0.005 & 0.660$\pm$0.005 & 0.657$\pm$0.006 & 0.653$\pm$0.003 & 0.651$\pm$0.002 & 0.653$\pm$0.003 \\
                           & Graph Classification & 0.342$\pm$0.053 & 0.358$\pm$0.029   & 0.325$\pm$0.029 & -               & 0.327$\pm$0.067 & 0.333$\pm$0.026 & 0.291$\pm$0.033 & -               \\ \midrule
\multirow{3}{*}{PRROTEINS} & Node Classification  & 0.916$\pm$0.017 & 0.925$\pm$0.017   & 0.916$\pm$0.017 & -               & 0.890$\pm$0.036 & 0.897$\pm$0.036 & 0.887$\pm$0.036 & -               \\
                           & Link Prediction      & 0.988$\pm$0.002 & 0.988$\pm$0.002   & 0.988$\pm$0.001 & 0.988$\pm$0.001 & 0.986$\pm$0.002 & 0.986$\pm$0.002 & 0.986$\pm$0.002 & 0.986$\pm$0.002 \\
                           & Graph Classification & 0.767$\pm$0.006 & 0.766$\pm$0.0.007 & 0.765$\pm$0.006 & -               & 0.759$\pm$0.005 & 0.757$\pm$0.003 & 0.757$\pm$0.003 & -               \\ \midrule
\multirow{3}{*}{Synthie}   & Node Classification  & -               & -                 & -               & -               & -               & -               & -               & -               \\
                           & Link Prediction      & 0.523$\pm$0.002 & 0.522$\pm$0.002   & 0.522$\pm$0.002 & 0.522$\pm$0.002 & 0.514$\pm$0.004 & 0.514$\pm$0.004 & 0.514$\pm$0.004 & 0.514$\pm$0.004 \\
                           & Graph Classification & 0.521$\pm$0.042 & 0.571$\pm$0.032   & 0.556$\pm$0.026 & -               & 0.535$\pm$0.016 & 0.532$\pm$0.017 & 0.522$\pm$0.022 & -               \\ \midrule
\multirow{3}{*}{ADIS}      & Node Classification  & 0.988$\pm$0.002 & 0.991$\pm$0.001   & 0.988$\pm$0.002 & -               & 0.985$\pm$0.005 & 0.989$\pm$0.003 & 0.986$\pm$0.003 & -               \\
                           & Link Prediction      & 0.933$\pm$0.002 & 0.927$\pm$0.003   & 0.927$\pm$0.003 & 0.927$\pm$0.003 & 0.931$\pm$0.001 & 0.926$\pm$0.002 & 0.925$\pm$0.002 & 0.926$\pm$0.002 \\
                           & Graph Classification & 0.984$\pm$0.002 & 0.984$\pm$0.002   & 0.984$\pm$0.002 & -               & 0.971$\pm$0.019 & 0.971$\pm$0.018 & 0.971$\pm$0.019 & -               \\ \bottomrule\bottomrule
\end{tabular}
\label{table:gnn}
\end{table}}
\begin{figure}[h]
    \centering
    \subfigure[Node Classification]{\includegraphics[width=0.33\textwidth]{{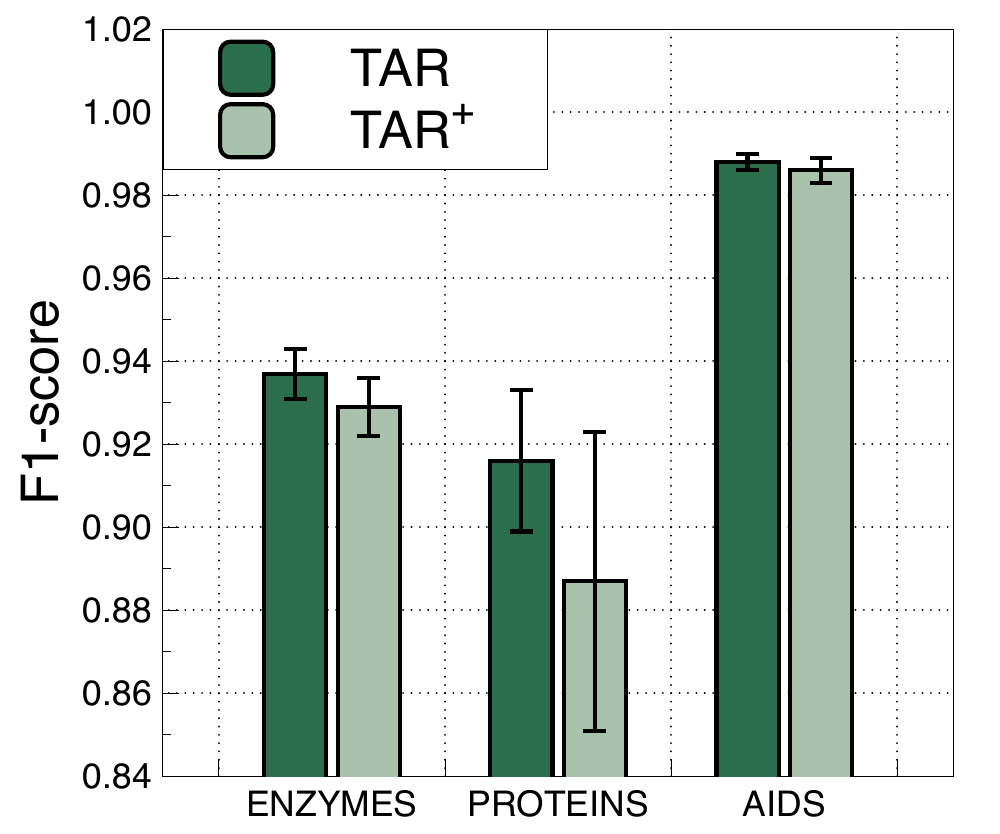}}}
    \subfigure[Link Prediction]{\includegraphics[width=0.33\textwidth]{{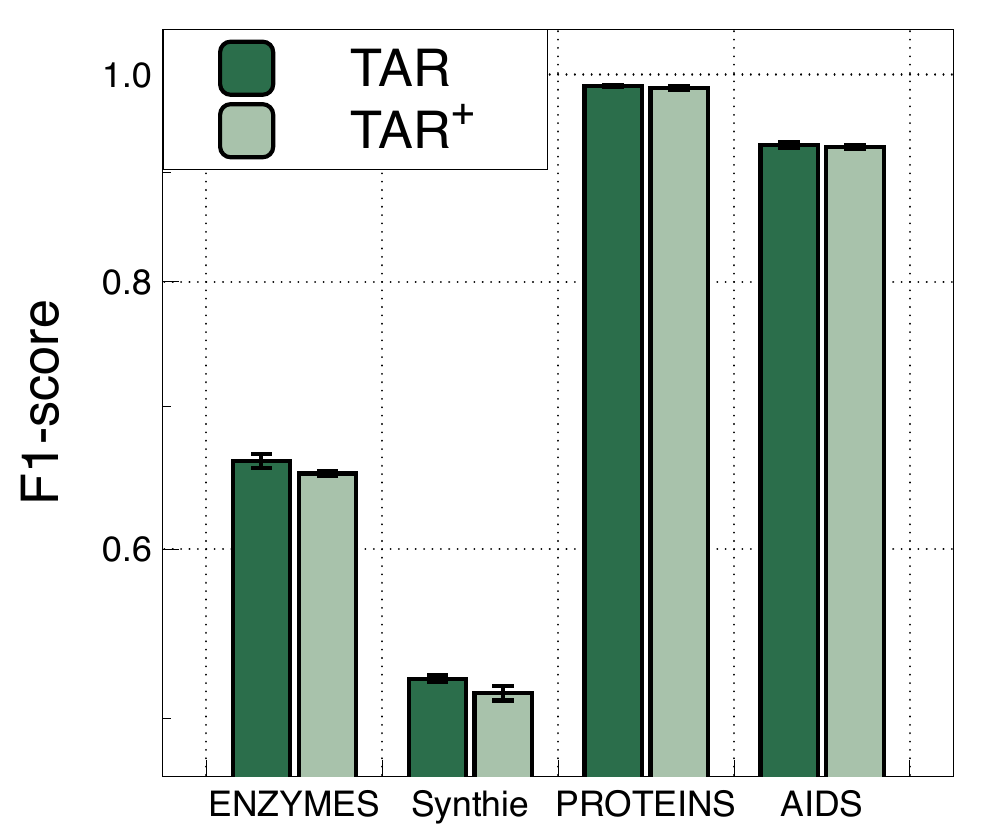}}}
    \subfigure[Graph Classification]{\includegraphics[width=0.33\textwidth]{{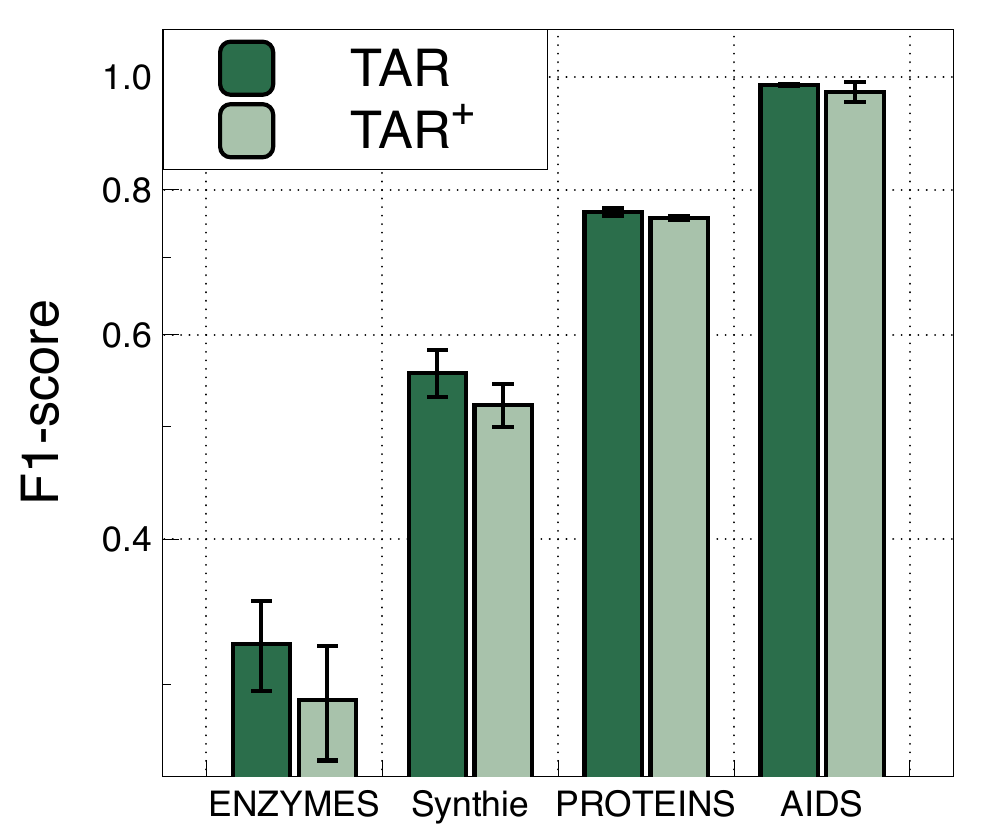}}}
    \caption{The impact of GNN-based state. F1-score comparisons of three different tasks across all datasets. We report mean values with standard deviation by 10 repeats.}
    \label{impact-gnn}
\end{figure}
\subsection{Impact of the GNN-based State (RQ3)}
In our framework, to capture the structure information of graphs, we adopt a well-trained GNN to embed a graph as a vector by aggregating the neighbors of nodes. Then we input the vector into the reinforcement agents so that the agents can make selection decisions based on the current state. To verify the impact of this component, we develop a variant model {\model$^+$}, which replaces the GNN with a descriptive statistics-based dual summarization method~\cite{dualsummar}. Figure~\ref{impact-gnn} demonstrates that {\model} outperforms {\model$^+$} in most cases. The underlying driver for this observation is that {\model} uses the GNN to be aware of the topological knowledge in graphs, enabling the model to consider this valuable structure information to reconstruct a better feature space. In conclusion, this experiment shows the importance of considering topological information when reconstructing a feature space for graph datasets. The detailed experiment results for all datasets are reported in Table~\ref{table:gnn}.

\setlength{\tabcolsep}{2.1mm}{
\begin{table}[h]
\centering
\fontsize{6}{9}\selectfont
\caption{The impact of grouping. Mean values with standard deviation by 10 repeats.}
\begin{tabular}{@{}c|c|cccc|cccc@{}}
\toprule\toprule
Dataset                    & Downstream Task      & \multicolumn{4}{c|}{\model}                                              & \multicolumn{4}{c}{\model$^*$}                          \\ \midrule
                           &                      & Precision       & Recall            & F1              & AUC             & Precision       & Recall          & F1              & AUC             \\ \midrule
\multirow{3}{*}{ENZYMES}   & Node Classification  & 0.936$\pm$0.006 & 0.943$\pm$0.005   & 0.937$\pm$0.006 & -               & 0.926$\pm$0.003 & 0.933$\pm$0.002 & 0.925$\pm$0.002 & -               \\
                           & Link Prediction      & 0.662$\pm$0.005 & 0.660$\pm$0.005   & 0.660$\pm$0.005 & 0.660$\pm$0.005 & 0.666$\pm$0.002 & 0.657$\pm$0.004 & 0.653$\pm$0.005 & 0.657$\pm$0.004 \\
                           & Graph Classification & 0.342$\pm$0.053 & 0.358$\pm$0.029   & 0.325$\pm$0.029 & -               & 0.339$\pm$0.040 & 0.348$\pm$0.003 & 0.306$\pm$0.004 & -               \\ \midrule
\multirow{3}{*}{PRROTEINS} & Node Classification  & 0.916$\pm$0.017 & 0.925$\pm$0.017   & 0.916$\pm$0.017 & -               & 0.893$\pm$0.004 & 0.898$\pm$0.006 & 0.886$\pm$0.007 & -               \\
                           & Link Prediction      & 0.988$\pm$0.002 & 0.988$\pm$0.002   & 0.988$\pm$0.001 & 0.988$\pm$0.001 & 0.987$\pm$0.001 & 0.986$\pm$0.001 & 0.986$\pm$0.001 & 0.986$\pm$0.001 \\
                           & Graph Classification & 0.767$\pm$0.006 & 0.766$\pm$0.0.007 & 0.765$\pm$0.006 & -               & 0.761$\pm$0.010 & 0.759$\pm$0.007 & 0.759$\pm$0.008 & -               \\ \midrule
\multirow{3}{*}{Synthie}   & Node Classification  & -               & -                 & -               & -               & -               & -               & -               & -               \\
                           & Link Prediction      & 0.523$\pm$0.002 & 0.522$\pm$0.002   & 0.522$\pm$0.002 & 0.522$\pm$0.002 & 0.519$\pm$0.003 & 0.519$\pm$0.003 & 0.519$\pm$0.003 & 0.519$\pm$0.003 \\
                           & Graph Classification & 0.521$\pm$0.042 & 0.571$\pm$0.032   & 0.556$\pm$0.026 & -               & 0.516$\pm$0.048 & 0.547$\pm$0.048 & 0.546$\pm$0.048 & -               \\ \midrule
\multirow{3}{*}{ADIS}      & Node Classification  & 0.988$\pm$0.002 & 0.991$\pm$0.001   & 0.988$\pm$0.002 & -               & 0.967$\pm$0.016 & 0.974$\pm$0.013 & 0.969$\pm$0.015 & -               \\
                           & Link Prediction      & 0.933$\pm$0.002 & 0.927$\pm$0.003   & 0.927$\pm$0.003 & 0.927$\pm$0.003 & 0.931$\pm$0.003 & 0.925$\pm$0.003 & 0.925$\pm$0.003 & 0.925$\pm$0.003 \\
                           & Graph Classification & 0.984$\pm$0.002 & 0.984$\pm$0.002   & 0.984$\pm$0.002 & -               & 0.965$\pm$0.030 & 0.966$\pm$0.029 & 0.965$\pm$0.031 & -               \\ \bottomrule\bottomrule
\end{tabular}
\label{table:grouping}
\end{table}}
\begin{figure}[h] 
    \centering
    \subfigure[Node Classification]{\includegraphics[width=0.33\textwidth]{{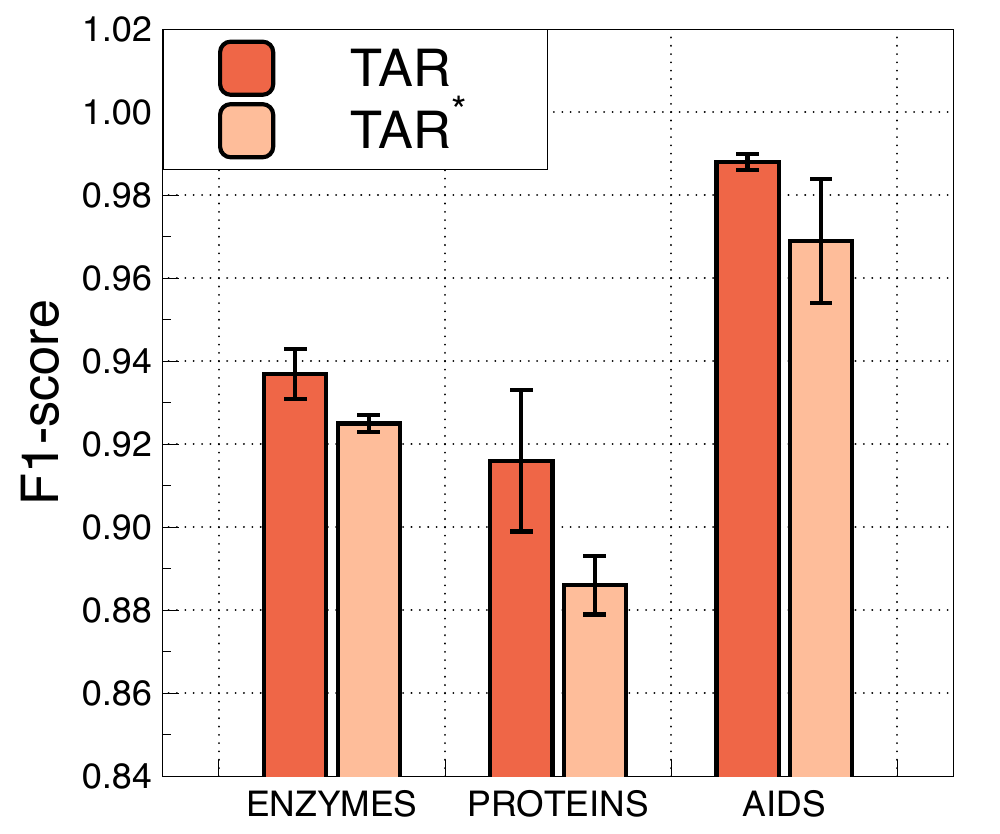}}}
    \subfigure[Link Prediction]{\includegraphics[width=0.33\textwidth]{{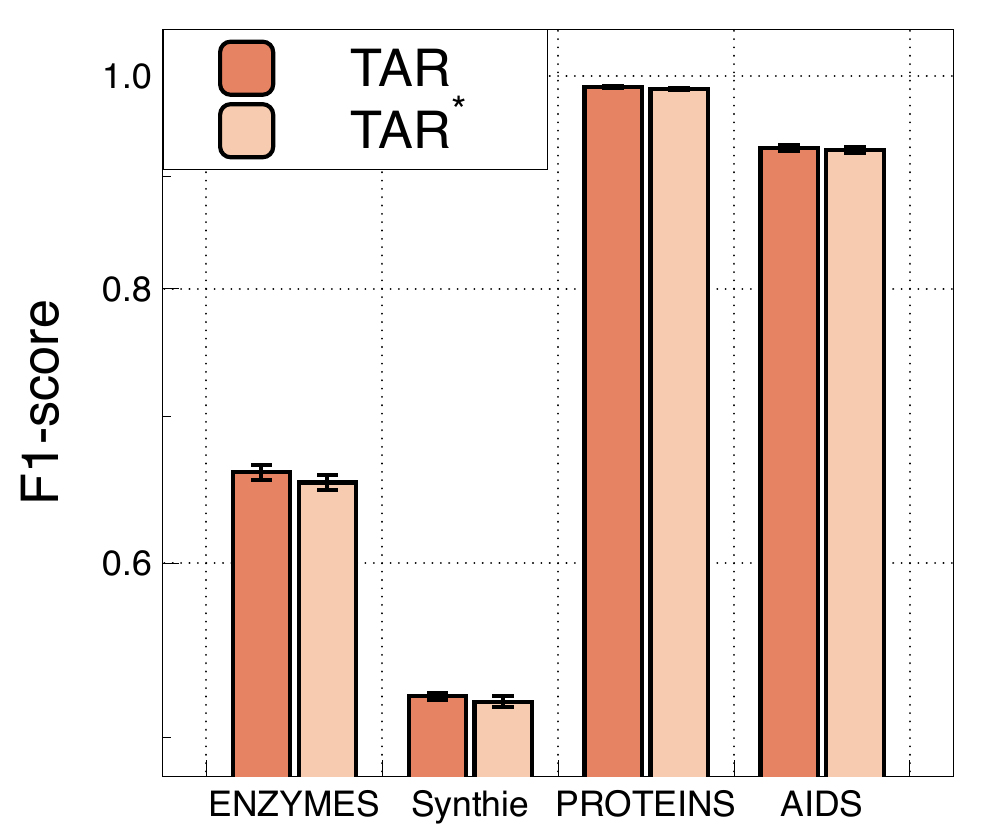}}}
    \subfigure[Graph Classification]{\includegraphics[width=0.33\textwidth]{{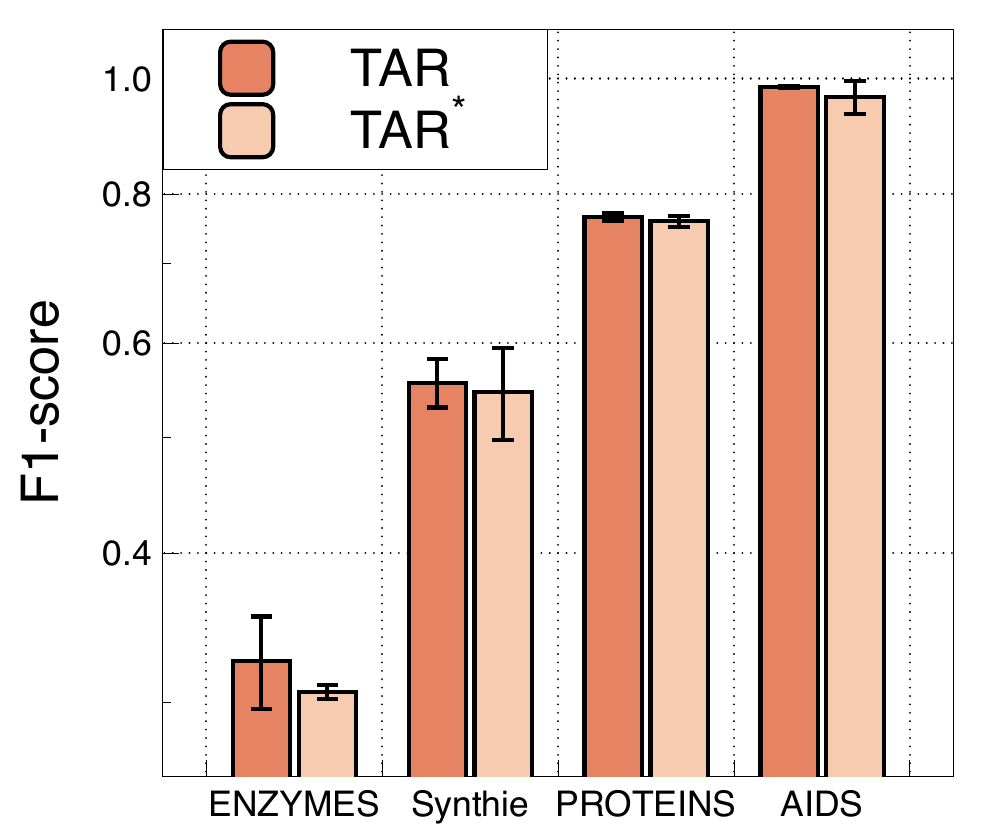}}}
    \caption{The impact of grouping. F1-score comparisons of three different tasks across all datasets. We report mean values with standard deviation by 10 repeats.}
    \label{impact-grouping}
\end{figure}

\subsection{Impact of the Feature Grouping (RQ4)}
To further analyze the impact of feature grouping, we removed this component, allowing the reinforcement agents to select an individual feature rather than a group of features within each step, denoted as {\model$^*$}. Figure~\ref{impact-grouping} indicates that {\model} surpasses {\model$^*$} across all datasets. This is attributed to {\model} generating more informative features and providing stronger feedback within each iteration. Additionally, {\model} generates a greater number of features compared to {\model$^*$}, which enhances the exploration of high-order crossed features within fixed steps. Therefore, this experiment underscores the necessity and effectiveness of feature grouping. The detailed experiment results for all datasets are reported in Table~\ref{table:grouping}.

\setlength{\tabcolsep}{4mm}{
\begin{table}[h]
\centering
\fontsize{6}{9}\selectfont
\caption{Time cost (s) between {\model} and baselines. Mean values with standard deviation by 10 repeats.}
\begin{tabular}{@{}c|c|ccccccc@{}}
\toprule\toprule
Dataset                   & Downstream Task      & PCA     & LDA        & RDG        & ERG          & NFS         & TTG         & {\model}        \\ \midrule
\multirow{3}{*}{ENZYMES}  & Node Classification  & 5$\pm$1 & 14$\pm$1   & 121$\pm$26 & 586$\pm$138  & 234$\pm$27  & 318$\pm$48  & 279$\pm$62 \\
                          & Link Prediction      & 5$\pm$1 & 14$\pm$0   & 126$\pm$20 & 611$\pm$171  & 237$\pm$66  & 393$\pm$113 & 271$\pm$66 \\
                          & Graph Classification & 3$\pm$1 & 18$\pm$10  & 63$\pm$16  & 419$\pm$9    & 134$\pm$38  & 237$\pm$20  & 148$\pm$26 \\ \midrule
\multirow{3}{*}{PROTEINS} & Node Classification  & 7$\pm$3 & 84$\pm$38  & 345$\pm$64 & 1944$\pm$444 & 577$\pm$128 & 575$\pm$170 & 385$\pm$47 \\
                          & Link Prediction      & 5$\pm$1 & 38$\pm$7   & 213$\pm$72 & 1963$\pm$458 & 254$\pm$28  & 903$\pm$212 & 313$\pm$22 \\
                          & Graph Classification & 4$\pm$1 & 67$\pm$35  & 135$\pm$46 & 1703$\pm$220 & 291$\pm$71  & 468$\pm$82  & 258$\pm$28 \\ \midrule
\multirow{2}{*}{Synthie}  & Link Prediction      & 5$\pm$1 & 112$\pm$41 & 197$\pm$48 & 768$\pm$174  & 283$\pm$59  & 427$\pm$107 & 397$\pm$93 \\
                          & Graph Classification & 3$\pm$1 & 110$\pm$42 & 48$\pm$14  & 731$\pm$162  & 158$\pm$35  & 357$\pm$69  & 278$\pm$40 \\ \midrule
\multirow{3}{*}{AIDS}     & Node Classification  & 5$\pm$1 & 44$\pm$22  & 168$\pm$24 & 183$\pm$4    & 234$\pm$63  & 250$\pm$55  & 264$\pm$77 \\
                          & Link Prediction      & 6$\pm$0 & 32$\pm$18  & 142$\pm$36 & 162$\pm$48   & 136$\pm$19  & 160$\pm$48  & 165$\pm$61 \\
                          & Graph Classification & 5$\pm$1 & 33$\pm$17  & 211$\pm$15 & 265$\pm$78   & 237$\pm$39  & 290$\pm$70  & 249$\pm$49 \\ \bottomrule\bottomrule
\end{tabular}
\label{table:complexity}
\end{table}}
\begin{figure}[h]
    \centering
    \subfigure[Node Classification]{\includegraphics[width=0.33\textwidth]{{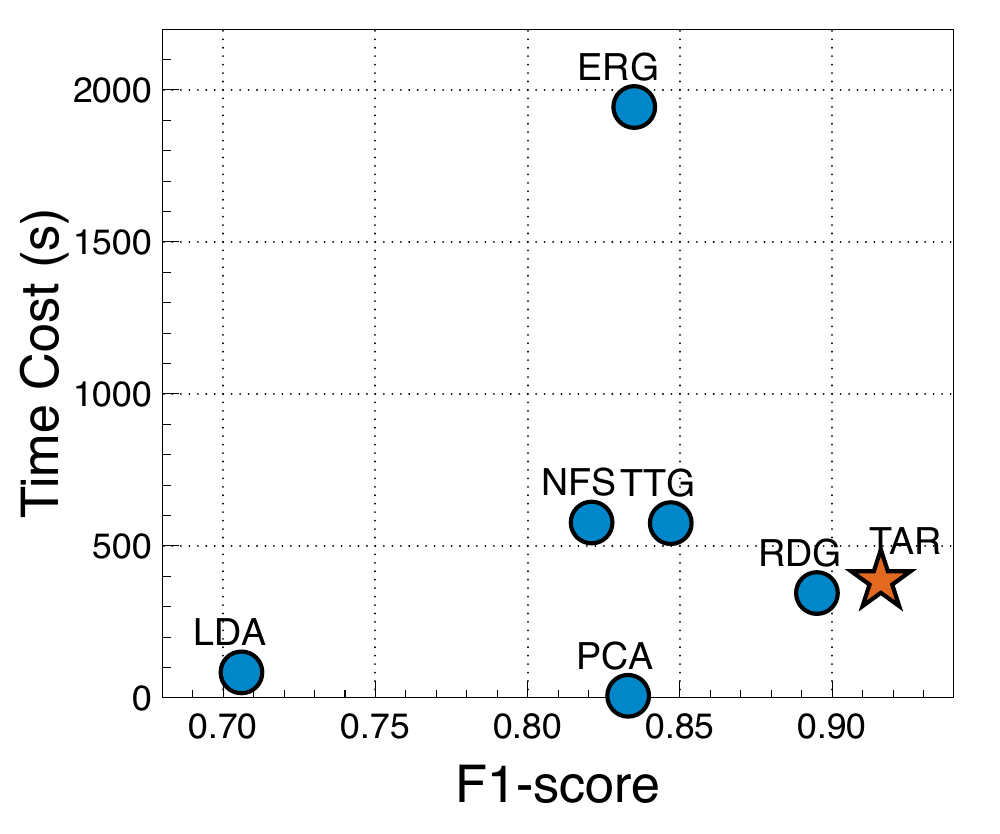}}}
    \subfigure[Link Prediction]{\includegraphics[width=0.33\textwidth]{{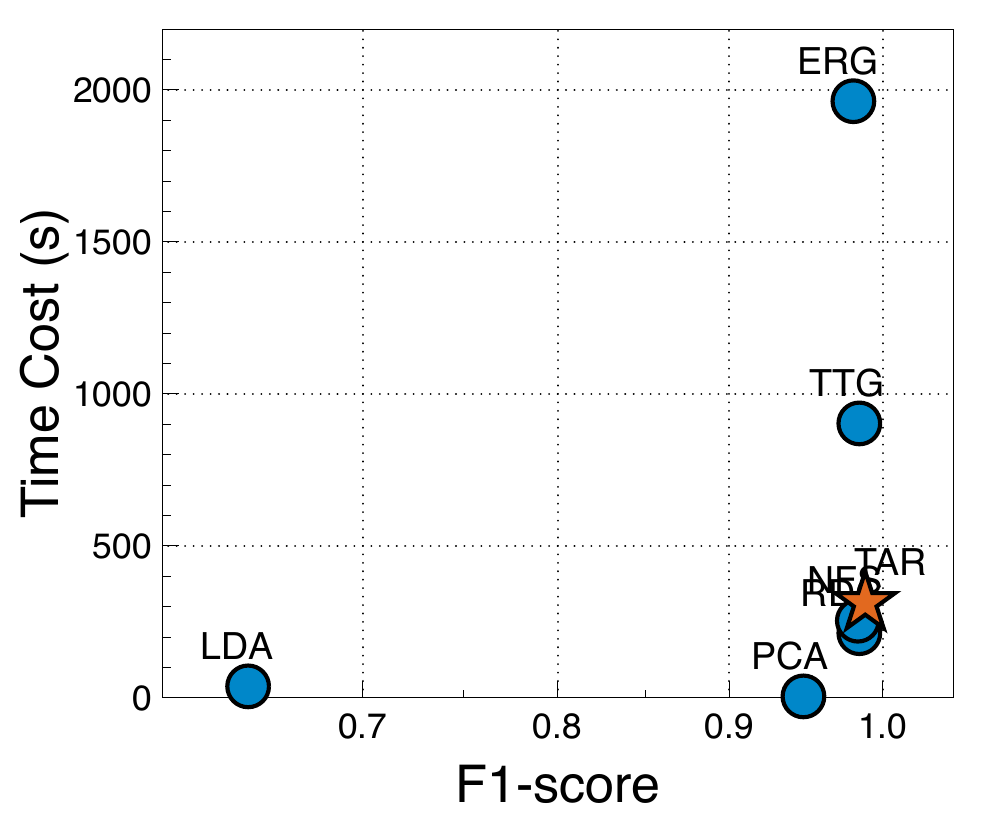}}}
    \subfigure[Graph Classification]{\includegraphics[width=0.33\textwidth]{{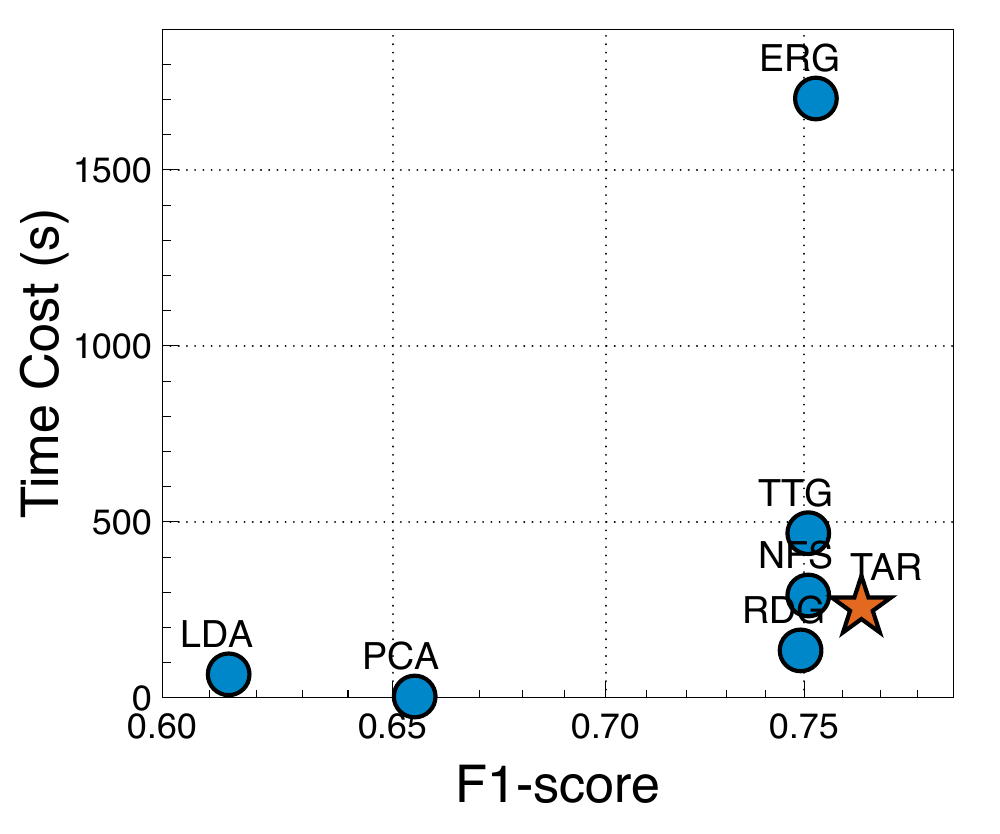}}}
    \caption{Complexity analysis on PROTEINS dataset. The x-axis and y-axis represent the F1-score and time cost corresponding to the specific downstream tasks, respectively.}
    \label{time-check}
\end{figure}
\subsection{Complexity Analysis (RQ5)}
We compared the computational complexity between {\model} and all baselines, as shown in Figure~\ref{time-check}. The x-axis illustrates the F1-score achieved in the downstream task of node classification using the PROTEINS dataset, while the y-axis represents the time cost required by each method. The unsupervised baselines, such as PAC and LDA, cost less time, but are less accurate. Compared with the supervised baselines, {\model} costs less time than ERG, NFS, and TTG, and costs a little more time than RDG. Remarkably, despite marginally higher time requirements compared to three baseline methods, {\model} achieves superior downstream performance, underscoring its ability to balance the effectiveness and efficiency of graph feature reconstruction. We report the complete results for all datasets in Table~\ref{table:complexity}.

\subsection{Reconstructed Feature Space Analysis (RQ6)}
We select the AIDS dataset as an example to visualize the reconstructed graph-level feature space. In detail, for a specific graph, we calculate the mean of all node embeddings to represent the graph. Then, we use T-SNE~\cite{tsne} to map them into a 2-dimensional space for visualization. 
\begin{figure}
    \centering
    \subfigure[Original]{\includegraphics[width=0.4\textwidth]{{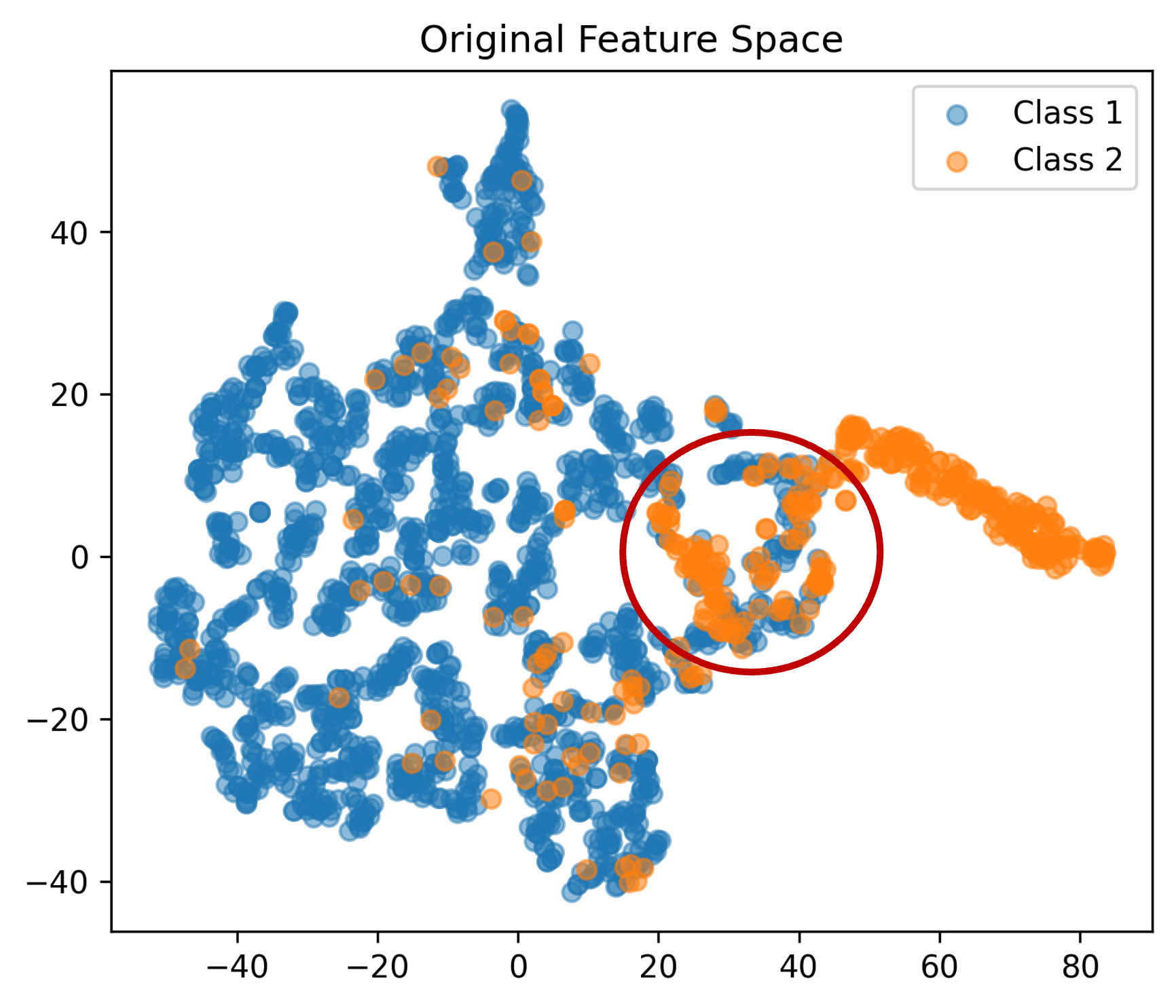}}}
    \subfigure[Reconstructed]{\includegraphics[width=0.4\textwidth]{{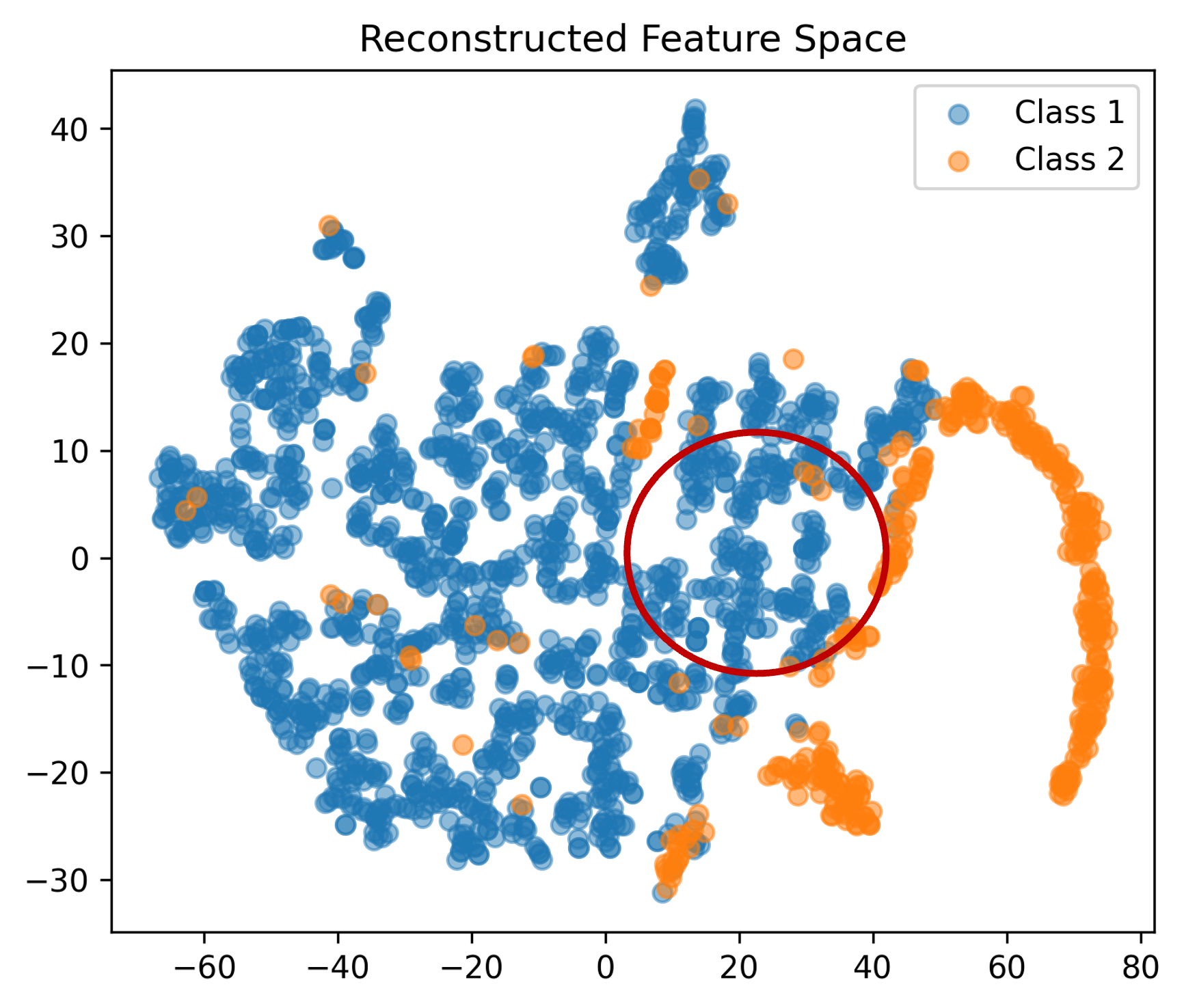}}}
    \caption{Visualization comparison between original reconstructed graph-level feature space for AIDS.}
    \label{case-visual}
\end{figure}
Figure~\ref{case-visual} shows the visualization results of the original feature space and reconstructed feature space, in which each point represents a unique graph in the AIDS dataset.
We can observe that graphs have a better distinctness in the reconstructed feature space. Especially for the graphs in the circle, {\model} can divide the graphs into two groups with a clear boundary. As a result, the F1-score of graph classification increases from 0.864 to 0.984, which is a significant improvement. The underlying driver is that {\model} can reconstruct the feature space aware of the graph structure information to make the dataset more discriminative, thus achieving better downstream performance.
\section{Related Work}

\textbf{Reinforcement Learning (RL)} is a machine learning method of how an intelligent agent ought to take actions in a dynamic environment in order to maximize the cumulative reward~\cite{RLintro}. The RL algorithms can be divided into two categories according to the learned policy: value-based and policy-based. Valued-based algorithm~\cite{valueRL1,valueRL2,zhang-etal-2024-prototypical} estimates the value function or action-value function to learn a policy. Policy-based~\cite{policyRL} learns a probability distribution to map state to action directly. Furthermore, an actor-critic reinforcement learning framework is introduced to integrate the benefits of both value-based and policy-based algorithms~\cite{bothRL}.

\noindent\textbf{Feature Space Reconstruction} aims to derive a new feature set from original features to improve downstream ML tasks~\cite{AFT1,AFT2,wabngyang@bio,wangyang@polymer}. Humans can manually reconstruct a feature space with domain knowledge and empirical experiences, which is explicit, traceable, and explainable, but incomplete and time-consuming. Machine-assistant methods can be divided into two categories: unsupervised and supervised methods. Unsupervised methods don't rely on labels such as Principal Components Analysis (PCA)~\cite{uddin2021pca}. PCA is based on a strong assumption of straight linear feature correlation, and only uses addition or subtraction to generate features, leading to only dimensionality reduction rather than an increase. Supervised methods include exhaustive-expansion-reduction approaches~\cite{expansion-reduction1,expansion-reduction2,expansion-reduction3,expansion-reduction4,expansion-reduction5}, iterative-feedback-improvement approaches~\cite{envolution2,envolution3,envolution4}, feature engineering with language models~\cite{gong2024evolutionarylargelanguagemodel,zhang2024ratt}, and AutoML techniques~\cite{automl4,automl5,automl6,zhang2024dynamicadaptivefeaturegeneration,zhang2024thoughtspaceexplorernavigating} that are enabled by reinforcement learning, genetic algorithms, or evolutionary algorithms. However, among a variety of data types~\cite{xie2024spatio}, all the methods are designed for tabular data, which makes it hard to generalize to graph datasets. In this paper, we develop a topological-aware reinforcement feature space reconstruction method, enabling a better feature space reconstruction for graph datasets.

\section{Conclusion Remarks}

In conclusion, our novel framework, which combines graph neural networks and topology-aware reinforcement learning, captures the complex topological structures in graph data to automatically reconstruct the optimal feature space, thereby improving downstream ML tasks.
Specifically, our approach captures significant structural details by mining core subgraphs and embedding their information with GNN. Then reinforcement learning agents iteratively generate features, reducing reliance on human intuition. This automated process achieves dual objectives: capturing graph topologies and optimizing feature generation.
This framework simplifies feature space reconstruction and improves the performance of machine learning models on graph data.
Extensive experiments validate the effectiveness and efficiency of including topological awareness. The superior experimental performance highlights its potential as a valuable approach for graph feature space reconstruction in large quantities of domains, such as bioinformatics, chemistry, etc.

\bibliographystyle{ACM-Reference-Format}
\bibliography{tkdd}

\end{document}